\documentclass[twoside]{article}

\usepackage[accepted]{aistats2023}

\usepackage{microtype}
\usepackage{graphicx}
\usepackage{subfigure}
\usepackage{booktabs} 
\usepackage{algorithm}
\usepackage[noend]{algpseudocode}

\usepackage{lipsum}
\usepackage{epsfig,epsf,fancybox}
\usepackage{amsmath}
\usepackage{mathrsfs}
\usepackage{amssymb}
\usepackage{amsfonts}
\usepackage{graphicx}
\usepackage{color}
\usepackage[linktocpage,pagebackref,colorlinks,linkcolor=blue,anchorcolor=blue,citecolor=blue,urlcolor=blue,hypertexnames=false]{hyperref}
\usepackage{stmaryrd}
\usepackage{multirow}
\usepackage{booktabs}
\usepackage{cite}
\usepackage{float}
\usepackage{bbm}
\usepackage[round]{natbib}

\usepackage{mathtools}

\usepackage[normalem]{ulem} 

\newcommand{\bm}[1]{\boldsymbol{#1}}

\newcommand{\epsoracle}{\epsilon_{\mathcal{A}}}
\newcommand{\Pc}{{\mathcal{P}}}
\newcommand{\valpha}{{\boldsymbol{\alpha}}}
\newcommand{\balpha}{{\boldsymbol{\alpha}}}

\newcommand{\vg}{{\mathbf{g}}}

\newcommand{\vu}{{\mathbf{u}}}
\newcommand{\vv}{{\mathbf{v}}}
\newcommand{\bu}{{\mathbf{u}}}
\newcommand{\bv}{{\mathbf{v}}}
\newcommand{\vw}{{\mathbf{w}}}
\newcommand{\bw}{{\mathbf{w}}}
\newcommand{\vx}{{\mathbf{x}}}
\newcommand{\bx}{{\mathbf{x}}}
\newcommand{\vy}{{\mathbf{y}}}
\newcommand{\by}{{\mathbf{y}}}
\newcommand{\vz}{{\mathbf{z}}}

\newcommand{\vF}{{\mathbf{F}}}
\newcommand{\vG}{{\mathbf{G}}}

\newcommand{\vxi}{{\bm{\xi}}}
\newcommand{\bxi}{{\bm{\xi}}}

\newcommand{\E}{\mathbb{E}}

\DeclareMathOperator*{\argmin}{arg\,min} 

\newcommand{\bc}{\begin{center}}
\newcommand{\ec}{\end{center}}

\newcommand{\bdm}{\begin{displaymath}}
\newcommand{\edm}{\end{displaymath}}

\newcommand{\beq}{\begin{equation}}
\newcommand{\eeq}{\end{equation}}

\newcommand{\bfl}{\begin{flushleft}}
\newcommand{\efl}{\end{flushleft}}

\newcommand{\bt}{\begin{tabbing}}
\newcommand{\et}{\end{tabbing}}

\newcommand{\beqn}{\begin{eqnarray}}
\newcommand{\eeqn}{\end{eqnarray}}

\newcommand{\beqs}{\begin{align*}} 
\newcommand{\eeqs}{\end{align*}}  

\newtheorem{assumption}{Assumption}

\newtheorem{theorem}{Theorem}

\newtheorem{lemma}{Lemma}
\newtheorem{definition}{Definition}
\newtheorem{proposition}{Proposition}
\newtheorem{example}{Example}
\newtheorem{remark}{Remark}
\begin{document}

%

%

\twocolumn[

\aistatstitle{Stochastic Methods for AUC Optimization subject to AUC-based Fairness Constraints}

\aistatsauthor{ Yao Yao \And Qihang Lin \And  Tianbao Yang }

\aistatsaddress{ University of Iowa \And  University of Iowa \And Texas A\&M University } ]

\begin{abstract}
As machine learning being used increasingly in making high-stakes decisions, an arising challenge is to avoid unfair AI systems that lead to discriminatory decisions for protected population. A direct approach for obtaining a fair predictive model is to train the model through optimizing its prediction performance subject to fairness constraints. Among various fairness constraints, the ones based on the area under the ROC curve (AUC) are emerging recently because they are threshold-agnostic and effective for unbalanced data. In this work, we formulate the problem of training a fairness-aware predictive model as an AUC optimization problem subject to a class of AUC-based fairness constraints. This problem can be reformulated as a min-max optimization problem with min-max constraints, which we solve by stochastic first-order methods based on a new Bregman divergence designed for the special structure of the problem. We numerically demonstrate the effectiveness of our approach on real-world data under different fairness metrics.  
\end{abstract}

\section{INTRODUCTION}
\label{sec:intro}
AI systems have been increasingly used to assist in making high-stakes decisions such as lending decision~\citep{addo2018credit}, bail and parole decision~\citep{dressel2018accuracy}, resource allocation~\citep{davahli2021optimizing} and so on. Along with this trend, a question arising is how to ensure AI systems are fair and do not produce discriminatory decisions for protected groups defined by some sensitive variables (e.g., age, race, gender, etc.). To answer this question, the first step is to define and quantitatively measure fairness of AI systems, which is itself an active research area. 

For a classification task, a variety of fairness metrics have been studied including demographic parity~\citep{calders2009building,beutel2019putting,gajane2017formalizing}, equality of opportunity~\citep{hardt2016equality}, equality of odds~\citep{hardt2016equality}, predictive quality parity~\citep{chouldechova2017fair} and counter factual fairness~\citep{kusner2017counterfactual}.  All of these fairness metrics are formulated based on statistical relationships between predicted class labels and sensitive variables. However, many predictive models only generate a predicted risk score and a predicted class label is obtained afterwards by comparing the score with a threshold. A good threshold is not always easy to choose in practice and may vary with datasets and applications. In fact, it is likely that a model satisfies a fairness criterion with one threshold but violates the same fairness criterion with another threshold. Moreover, the threshold is often chosen to achieve a targeted predicted positive rate. When doing so, it is not easy to ensure a targeted fairness criterion is satisfied at the same threshold.


With these drawbacks in the threshold-dependent fairness metrics, there have been growing efforts on developing threshold independent fairness metrics, among which the fairness metrics based on AUC, or equivalently, pairwise comparison, are prevalent~\citep{kallus2019fairness, dixon2018measuring,narasimhan2020pairwise,borkan2019nuanced,beutel2019fairness,vogel2021learning,yang2022minimax}. These  metrics are directly defined based on statistical relationships between predicted risk scores and sensitive variables and thus do not require a predetermined threshold.

Regardless of the fairness metric applied, training a fair predictive model requires balancing the model's prediction performance and fairness, two potentially conflicting targets. Hence, it is naturally to formulate this problem as constrained optimization where the model's prediction performance is optimized subject to some fairness constraints. This approach has been studied with constraints based on threshold-dependent fairness metrics~\citep{agarwal2018reductions,kearns2018preventing,woodworth2017learning,goh2016satisfying,diana2021minimax,dwork2012fairness,cotter2019optimization,cruz2022fairgbm,cotter2018training} and threshold-agnostic fairness metrics~\citep{vogel2021learning,narasimhan2020pairwise,zafar2017} with different optimization algorithms applied during training. In~\cite{narasimhan2020pairwise}, a proxy-Lagrangian method from \cite{cotter2019optimization,cotter2018training} is applied to optimization with fairness constraints while regularization methods are applied by \cite{vogel2021learning,beutel2017data} to optimize a weighted sum of prediction performance and fairness metrics.

Online learning is a common setting  in machine learning where data becomes available sequentially and the model needs to be updated by the latest data. When learning a fair model online, the methods in~\cite{vogel2021learning,narasimhan2020pairwise} need to compute stochastic gradients of the constraint functions. However,  due to the pairwise comparison involved in their optimization models, computing one online stochastic gradient requires processing a pair of data points, one from the protected group and the other from the unprotected group. This requires that data points always arrive in pairs, which is not always guaranteed in practice. For the similar reason, when training models off-line using an existing dataset,  the methods by~\cite{vogel2021learning,narasimhan2020pairwise} require processing all pairs of data points and thus need a computational cost quadratic in data size, which is prohibited for large datasets. 

In this paper, we focus on developing efficient numerical methods for training a classification model under AUC-based threshold-agnostic fairness constraints by addressing the computational issues mentioned above. The main contribution of this paper is formulating the aforementioned problem into a stochastic optimization problem subject to min-max constraints. Although the min-max constraints are new and challenging structures, we propose a special Bregman divergence after changing variables such that the problem can be solved efficiently by the existing stochastic first-order methods for constrained stochastic optimization such as~\cite{lin2020data,boob2022stochastic,ma2020quadratically}. Compared to~\cite{vogel2021learning,narasimhan2020pairwise}, the main advantage of our approach is that it supports model training in an online setting with one data point, instead of a data pair, arriving each time in any sequence. Moreover, when applied under the off-line setting, our approach only has a computational cost linear in data size. One limitation of our approach is that we must use a quadratic surrogate loss to approximate the AUCs in the objective and constraint functions. However, the numerical results on real-world datasets show that the models found by our methods trade off classification performance and fairness more effectively than existing techniques.

\section{RELATED WORKS}

A survey of prevalent fairness metrics, including some discussed in the previous section, is provided by~\cite{verma2018fairness}. However, most metrics discussed in~\cite{verma2018fairness} are based on predicted class labels and thus threshold dependent. The threshold-agnostic fairness metrics based on AUC (see examples in Section~\ref{sec:preliminary}) have been proposed in~\cite{borkan2019nuanced,kallus2019fairness,dixon2018measuring,vogel2021learning}. They have been extended to a broader class of metrics based on pairwise comparison, so the target variable can be continuous or ordinal (e.g., in a regression or ranking problem)~\citep{narasimhan2020pairwise,beutel2019fairness}. The class of fairness metrics we consider in this paper is more general than~\cite{borkan2019nuanced,kallus2019fairness,dixon2018measuring} and has similar generality to~\cite{narasimhan2020pairwise,beutel2019fairness}. A ROC-based fairness metric is proposed by~\cite{vogel2021learning} which is threshold-agnostic and stronger than the AUC-based ones in this paper. However, their optimization algorithms do not have theoretical convergence guarantees and require processing data points in pairs per iteration, which leads to a quadratic computational cost and is not ideal for training online. 

The three main approaches for building a fairness-aware machine learning model include the pre-processing, post-processing, and in-processing methods. The pre-processing method reduce machine bias by re-sampling and balancing training data~\citep{dwork2012fairness}. The post-processing method adjusts the prediction results after to ensure fairness~\citep{hardt2016equality}. The methods in this paper are the in-processing methods, which enforce fairness of a model during training by adding constraints or regularization to the optimization problem~\citep{agarwal2018reductions,goh2016satisfying,yang2022minimax}. 

Most in-processing methods are based on threshold-dependent fairness metrics~\citep{agarwal2018reductions,kearns2018preventing,woodworth2017learning,goh2016satisfying,diana2021minimax,dwork2012fairness,cotter2019optimization,cruz2022fairgbm,cotter2018training} while this work considers threshold-independent metrics. The unconstrained optimization approach by~\cite{yang2022minimax} minimizes the maximum of four different AUC scores to achieve a balance between classification performance and  fairness, while we ensure fairness by constraints. Although a constrained optimization approach is also presented in the appendix of~\cite{yang2022minimax}, no convergence analysis is provided. Fairness constrained optimization is an important application of stochastic constrained optimization for which many effective algorithms have been developed under the convex setting~\citep{lin2020data,boob2022stochastic,yan2022adaptive,yang2022data} and the non-convex setting~\citep{boob2022stochastic,ma2020quadratically}.  A proxy Lagrangian method has been developed for optimization subject to a class of rate constraints~\citep{narasimhan2020pairwise,cotter2019optimization,cotter2018training}, which include almost all fairness constraints we discussed above. The theoretical complexity of the proxy Lagrangian method has been analyzed~\citep{cotter2019optimization} when the objective function is convex or non-convex~\citep{cotter2019optimization} although a strong Bayesian optimization oracle is assumed in the non-convex case. Unconstrained optimization has also been considered for building fairness-aware models where fairness is enforced through a penalty term~\citep{beutel2017data,beutel2019fairness,vogel2021learning}. 

When directly applied to the AUC-based fairness constraints, the optimization algorithms mentioned above all need to request a pair of data points, one from the protected group and one from the opposite group, to construct the stochastic gradients. This is not ideal for online learning because data may not always arrive in pairs. On the contrary, our method is developed by first reformulating the AUC-based fairness constraints into min-max constraints using a quadratic loss~\citep{ying2016stochastic}. The stochastic gradient of this formulation can be computed using only one data point each time with any order of arrivals. Min-max stochastic constraints are new in optimization literature, so we develop a new Bregmen divergence by changing variables so that the existing algorithms like~\cite{lin2020data,boob2022stochastic} and their convergence analysis can be applied. 
\cite{yang2022stochastic} develop an algorithm for stochastic compositional optimization subject to compositional constraints which can be applied to our problem with the same computational complexity. This is because our min-max constraints can be also viewed as compositional constraints. They focus on the convex case but we also consider the non-convex case under some additional assumption (see Assumption~\ref{assum:wconvex}). 




\section{PRELIMINARIES}
\label{sec:preliminary}
Consider a binary classification problem, where the goal is to build a model that predicts a binary label $\zeta \in \{1,-1\}$ based on a feature vector $\vxi \in \mathbb{R}^p$. The sensitive feature of a data point is denoted by $\gamma \in \{1,-1\}$, which may or may not be a coordinate of $\vxi$. This feature divides the data into a protected group ($\gamma=1$) and an unprotected group ($\gamma=-1$). We denote a data point by a triplet 
$\vz= (\vxi,\zeta,\gamma)\in\mathbb{R}^{p+2}$ which is a random vector. We say $\mathcal{G}\subset\mathbb{R}^{p+2}$ has a \emph{positive measure w.r.t.} $\vz$ if $\Pr(\vz\in\mathcal{G})>0$. Let $h_{\vw}:\mathbb{R}^p\rightarrow \mathbb{R}$ be the predictive model that  produces a score $h_{\vw}(\vxi)$ for $\vxi$. Function $h_{\vw}$ is parameterized by a vector $\vw$ from a convex compact set $\mathcal{W}\subset\mathbb{R}^d$. We assume $h_{\vw}(\cdot)$ is differentiable and consider threshold-agnostic fairness metrics defined based on the join distribution of $h_{\vw}(\vxi)$, $\zeta$ and $\gamma$.

\begin{definition}[AUC defined by subsets]
\label{def:AUC}
Let $\vz= (\vxi,\zeta,\gamma)$ and $\vz' = (\vxi',\zeta',\gamma')$ be i.i.d random data points. Given two sets $\mathcal{G}$ and $\mathcal{G}'$ in $\mathbb{R}^{p+2}$ with positive measures w.r.t. $\vz$, the AUC w.r.t. $\mathcal{G}$ and $\mathcal{G}'$ is 
$$
\text{AUC}_\vw(\mathcal{G},\mathcal{G}'):=\text{Pr}(h_{\vw}(\vxi)>h_{\vw}(\vxi')|\vz\in\mathcal{G},\vz'\in\mathcal{G'}).
$$
\end{definition}
When $\mathcal{G}=\mathcal{D}_+:=\{\vz|\zeta=1\}$ and $\mathcal{G}'=\mathcal{D}_-:=\{\vz|\zeta=-1\}$, $\text{AUC}_\vw(\mathcal{G},\mathcal{G}')$ is reduced to the standard AUC for a binary classification problem. 

\begin{definition}[AUC-based fairness metric]
\label{def:AUCmetric}
Given sets $\mathcal{G}_1$, $\mathcal{G}_2$, $\mathcal{G}_1'$ and $\mathcal{G}_2'$ in $\mathbb{R}^{p+2}$ with positive measures w.r.t. $\vz$, the AUC-based fairness metric w.r.t. $\mathcal{G}_1$, $\mathcal{G}_2$, $\mathcal{G}_1'$ and $\mathcal{G}_2'$ is 
\begin{align}
\label{eq:AUCmetric}
|\text{AUC}_\vw(\mathcal{G}_1,\mathcal{G}_1')-\text{AUC}_\vw(\mathcal{G}_2,\mathcal{G}_2')|\in[0,1],
\end{align}
where $\text{AUC}_\vw(\cdot,\cdot)$ follows Definition~\ref{def:AUC}.
\end{definition}


We say model $h_{\vw}$ is unfair if the value of \eqref{eq:AUCmetric} is close to one and is fair if close to zero. This class of fairness metrics contains several existing threshold-agnostic metrics in literature, including the inter-group pairwise fairness~\citep{kallus2019fairness,beutel2019fairness}, the intra-group pairwise fairness~\citep{beutel2019fairness}, the positive/negative average equality gap~\citep{borkan2019nuanced} and the fairness metric based on background-subgroup AUCs~\citep{borkan2019nuanced}. In Appendix~\ref{sec:example}, we discuss how Definition~\ref{def:AUCmetric} is reduced to these metrics by setting  $\mathcal{G}_1$, $\mathcal{G}_2$, $\mathcal{G}_1'$ and $\mathcal{G}_2'$ to be different sets.

Besides fairness, we are also interested in the performance of the model as a classifier. In this paper, we also use the AUC, namely,  $\text{AUC}_\vw(\mathcal{D}_+,\mathcal{D}_-)$, as the performance metric and optimize it subject to fairness constraints. This choice is made only to obtain a uniform structure in the objective and constraint functions. The numerical methods we presented in this paper can be also applied when the classification performance is optimized by a traditional method, e.g., minimizing the empirical logistic loss.


The general formulation of our problem can be written as 
\begin{equation*}
\begin{split}
    \max_{\vw\in \mathcal{W}} \quad & \text{AUC}_\vw(\mathcal{D}_+,\mathcal{D}_-),\\
    \text{s.t.} \quad & |\text{AUC}_\vw(\mathcal{G}_1,\mathcal{G}_1')-\text{AUC}_\vw(\mathcal{G}_2,\mathcal{G}_2')|=0.
\end{split}
\end{equation*}
The equality constraint used here may be too restrict because an absolutely fair model may have a poor prediction performance and may be unnecessarily overly fair for users. To provide some flexibility to users, we replace the equality constraint to two inequalities after introducing a targeted \emph{level of fairness}, denoted by $\kappa\geq 0$, on the right-hand sides: 
\begin{equation}\label{eqn:problem}
\begin{split}
 \max_{\vw\in \mathcal{W}} \quad &\text{AUC}_\vw(\mathcal{D}_+,\mathcal{D}_-),\\
    \text{s.t.} \quad & \text{AUC}_\vw(\mathcal{G}_1,\mathcal{G}_1')-\text{AUC}_\vw(\mathcal{G}_2,\mathcal{G}_2') \leq \kappa,\\
    & \text{AUC}_\vw(\mathcal{G}_2,\mathcal{G}_2')-\text{AUC}_\vw(\mathcal{G}_1,\mathcal{G}_1')\leq \kappa.
\end{split}
\end{equation}


Solving \eqref{eqn:problem} directly is challenging because the objective and constraint functions involve indicator functions which are discontinuous. A common solution is to introduce a surrogate loss to approximate the indicator function. In particular, focusing on the objective function first, we have 
\begin{eqnarray}
\nonumber
&&\max_{\vw\in \mathcal{W}}  \text{AUC}_\vw(\mathcal{D}_+,\mathcal{D}_-)\\\nonumber
&=&\max_{\vw\in\mathcal{W}}\Pr(h_{\vw}(\vxi)>h_{\vw}(\vxi')|\zeta=1, \zeta'=-1 )\\\nonumber
&\Leftrightarrow&\min_{\vw\in \mathcal{W}}\Pr(h_{\vw}(\vxi)\leq h_{\vw}(\vxi')|\zeta=1, \zeta'=-1 )\\\nonumber
&=&\min_{\vw\in \mathcal{W}} \E\big[\mathbb{I}_{(h_{\vw}(\vxi)-h_{\vw}(\vxi')\leq 0)}|\zeta=1, \zeta'=-1\big] \\\label{eq:loss}
&\approx&\min_{\vw}\E\big[\ell(h_{\vw}(\vxi)-h_{\vw}(\vxi'))|\zeta=1, \zeta'=-1\big],
\end{eqnarray}
where $\ell(\cdot)$ is a continuous surrogate loss function that approximates the indicator functions $\mathbb{I}_{(\cdot\leq 0)}$ and $\mathbb{I}_{(\cdot< 0)}$. Similar to \eqref{eq:loss}, we approximate the left-hand side of the first constraint in \eqref{eqn:problem} as follows
\begin{align}
\nonumber
&\text{AUC}_\vw(\mathcal{G}_1,\mathcal{G}_1')-\text{AUC}_\vw(\mathcal{G}_2,\mathcal{G}_2') \\\nonumber
=&\text{Pr}(h_{\vw}(\vxi)>h_{\vw}(\vxi')|\vz\in\mathcal{G}_1,\vz'\in\mathcal{G}_1')\\\nonumber
&-\text{Pr}(h_{\vw}(\vxi)>h_{\vw}(\vxi')|\vz\in\mathcal{G}_2,\vz'\in\mathcal{G}_2') \\\nonumber
=&\text{Pr}(h_{\vw}(\vxi)>h_{\vw}(\vxi')|\vz\in\mathcal{G}_1,\vz'\in\mathcal{G}_1')\\\nonumber
&+\text{Pr}(h_{\vw}(\vxi)\leq h_{\vw}(\vxi')|\vz\in\mathcal{G}_2,\vz'\in\mathcal{G}_2') -1\\\nonumber
\approx&\E\big[\ell(h_{\vw}(\vxi')-h_{\vw}(\vxi))|\vz\in\mathcal{G}_1,\vz'\in\mathcal{G}_1'\big]\\\label{eq:constraint1}
&+\E\big[\ell(h_{\vw}(\vxi)-h_{\vw}(\vxi'))|\vz\in\mathcal{G}_2, \vz'\in\mathcal{G}_2'\big]-1.
\end{align}
Similarly, we approximate the left-hand side of the second constraint in \eqref{eqn:problem} as
\begin{align}
\nonumber
&\text{AUC}_\vw(\mathcal{G}_2,\mathcal{G}_2')-\text{AUC}_\vw(\mathcal{G}_1,\mathcal{G}_1') \\\nonumber
\approx&\E\big[\ell(h_{\vw}(\vxi')-h_{\vw}(\vxi))|\vz\in\mathcal{G}_2,\vz'\in\mathcal{G}_2'\big]\\\label{eq:constraint2}
&+\E\big[\ell(h_{\vw}(\vxi)-h_{\vw}(\vxi'))|\vz\in\mathcal{G}_1,\vz'\in\mathcal{G}_1'\big]-1.
\end{align}
Using \eqref{eq:loss} as the objective function and \eqref{eq:constraint1} and \eqref{eq:constraint2} as the left-hand sides of the inequality constraints. We obtain the following approximation to \eqref{eqn:problem}.
\begin{align}
\label{eq:approxproblem}
\min_{\vw\in\mathcal{W}}\ &\E\big[\ell(h_{\vw}(\vxi)-h_{\vw}(\vxi'))|\zeta=1, \zeta'=-1\big]\\\nonumber
\text{s.t. }& \E\big[\ell(h_{\vw}(\vxi')-h_{\vw}(\vxi))|\vz\in\mathcal{G}_1,\vz'\in\mathcal{G}_1'\big],\\\nonumber
&+\E\big[\ell(h_{\vw}(\vxi)-h_{\vw}(\vxi'))|\vz\in\mathcal{G}_2,\vz'\in\mathcal{G}_2'\big]\leq 1+\kappa,\\\nonumber
& \E\big[\ell(h_{\vw}(\vxi')-h_{\vw}(\vxi))|\vz\in\mathcal{G}_2,\vz'\in\mathcal{G}_2'\big]\\\nonumber
&+\E\big[\ell(h_{\vw}(\vxi)-h_{\vw}(\vxi'))|\vz\in\mathcal{G}_1,\vz\in\mathcal{G}_1'\big]\leq 1+\kappa.
\end{align}
Although \eqref{eq:approxproblem} have continuous objective and constraint functions, it is still computationally challenging in general because each expectation in \eqref{eq:approxproblem} is taken over a pair of random data points from two different subsets. When formulated using the empirical distribution over $n$ data points, each expectation becomes double summations which have $O(n^2)$ computational cost. Moreover, \eqref{eq:approxproblem} is not suitable for online learning as computing its stochastic gradient requires data arriving in pairs (one from $\mathcal{G}_i$ and one from $\mathcal{G}_i'$), which is not always the case. Fortunately, when the loss function is quadratic, more specifically, when $\ell(\cdot)=c_1(\cdot- c_2)^2$ with $c_1,c_2>0$, it is shown by~\cite{ying2016stochastic} that each expected loss in \eqref{eq:approxproblem} can be reformulated as the optimal value of a min-max optimization problem whose objective function can be computed in $O(n)$ cost under the empirical distribution. The new formulation also supports online learning since its stochastic gradient can be computed even with one data point (see Lemma~\ref{lemma1} below). To derive the reformulation of \eqref{eq:approxproblem} with quadratic loss functions, we need the following lemma by \citep{ying2016stochastic} whose proof is provided in Appendix~\ref{sec:lemma1} just for completeness.



\begin{lemma}
\label{lemma1}
Let $\vz = (\vxi,\zeta,\gamma)$ and $\vz' = (\vxi',\zeta',\gamma')$ be i.i.d random data points. Given any two sets $\mathcal{G}$ and $\mathcal{G}'$ in $\mathbb{R}^{p+2}$ with positive measures w.r.t. $\vz$,
\small
	\begin{equation}
		\begin{split}
		&\mathbb{E}\left[c_1( h_{\vw}(\vxi)-h_{\vw}(\vxi')-c_2 )^2 |\vz\in\mathcal{G},\vz'\in\mathcal{G}'\right]= \\\label{eq:minmax}
		& \hspace{-0.2cm}\min_{a,b\in \mathcal{I}_{\mathcal{G},\mathcal{G}'}}\max_{\alpha\in\mathcal{I}_{\mathcal{G},\mathcal{G}'}}\hspace{-0.2cm}\mathbb{E}\left\{ \hspace{-0.05cm}F_{\mathcal{G},\mathcal{G}'}(\vw,a,b;\vz)+\alpha G_{\mathcal{G},\mathcal{G}'}(\vw; \vz)-\alpha^2\hspace{-0.05cm}\right\},
		\end{split}
	\end{equation}
\normalsize
where
\small
\begin{equation}
\label{eq:GFDef}
	\begin{split}
&F_{\mathcal{G},\mathcal{G}'}(\vw,a,b;\vz):=\\
&\quad c_1c_2^2-\frac{2c_1c_2 h_{\vw}(\vxi)\mathbb{I}_{\mathcal{G}}(\vz) }{\text{Pr}(\vz\in\mathcal{G})}+\frac{2c_1c_2 h_{\vw}(\vxi)\mathbb{I}_{\mathcal{G}'}(\vz) }{\text{Pr}(\vz\in\mathcal{G}')}\\
&\quad +\frac{ c_1(h_{\vw}(\vxi)-a)^2\mathbb{I}_{\mathcal{G}}(\vz) }{\text{Pr}(\vz\in\mathcal{G})}+\frac{ c_1(h_{\vw}(\vxi)-b)^2\mathbb{I}_{\mathcal{G}'}(\vz) }{\text{Pr}(\vz\in\mathcal{G}')},\\
&G_{\mathcal{G},\mathcal{G}'}(\vw; \vz):=\frac{2c_1 h_{\vw}(\vxi)\mathbb{I}_{\mathcal{G}}(\vz) }{\text{Pr}(\vz\in\mathcal{G})}-\frac{2c_1 h_{\vw}(\vxi)\mathbb{I}_{\mathcal{G}'}(\vz) }{\text{Pr}(\vz\in\mathcal{G}')},
\end{split}
\end{equation} 
\normalsize
and $\mathcal{I}_{\mathcal{G},\mathcal{G}'}\subset\mathbb{R}$ is the smallest interval such that
\small
$$
\begin{array}{c}
0, \pm\E\big[h_{\vw}(\vxi)|\vz\in\mathcal{G}\big],\pm\E\big[h_{\vw}(\vxi')|\vz'\in\mathcal{G}'\big],\\
\pm\left(\E\big[h_{\vw}(\vxi)|\vz\in\mathcal{G}\big]-\E\big[h_{\vw}(\vxi')|\vz'\in\mathcal{G}'\big]\right)
\end{array}
\in \mathcal{I}_{\mathcal{G},\mathcal{G}'}
$$
\normalsize
for any $\vw\in\mathcal{W}$.
\end{lemma}

According to Lemma~\ref{lemma1}, the new formulation \eqref{eq:minmax} needs three auxiliary variables, $a$, $b$ and $\alpha$ in a large enough interval $\mathcal{I}_{\mathcal{G},\mathcal{G}'}$. We then apply Lemma~\ref{lemma1} to each conditional expected loss in~\eqref{eq:approxproblem} with $\ell(\cdot)=c_1(\cdot- c_2)^2$. To do so, we first define $\mathcal{I}$ as any bounded interval such that
\begin{equation}
\label{eq:intervalI}
\mathcal{I}_{\mathcal{D}_+, \mathcal{D}_-},\mathcal{I}_{\mathcal{G}_1, \mathcal{G}_1'},\mathcal{I}_{\mathcal{G}_2,\mathcal{G}_2'}\subset \mathcal{I},
\end{equation}
\normalsize
where $\mathcal{I}_{\mathcal{G},\mathcal{G}'}$ is defined as in Lemma~\ref{lemma1}. We then introduce fifteen auxiliary variables $a_i$, $b_i$ and $\alpha_i$ in $\mathcal{I}$ for $i=0,\dots,4$. Here, $(a_i,b_i,\alpha_i)$ for each $i$ corresponds to one  conditional expected loss in~\eqref{eq:approxproblem} (there are five of them). In additional, we define the primal variable $\vx=(\vw,(a_i,b_i)_{i=0}^{4})\in \mathcal{X}:=\mathcal{W}\times\mathcal{I}^{10}$ and the dual variable $\valpha=(\alpha_i)_{i=0}^4\in \mathcal{I}^{5}$. With these notations, we apply Lemma~\ref{lemma1} and reformulate~\eqref{eq:approxproblem} as  
\begin{align}
\label{eq:minmaxproblem}
\hspace{-0.2cm}f^*:=\min&f_0(\vx)~
\text{s.t.}~f_1(\vx)\leq 1+\kappa,\  f_2(\vx)\leq 1+\kappa,
\end{align}
where
\begin{small}
\begin{align}
\label{eq:f0}
\hspace{-0.2cm}f_0(\vx):=&\hspace{-0.05cm}\max_{\alpha_0\in\mathcal{I}} \hspace{-0.05cm} \mathbb{E}\hspace{-0.1cm} \left[\hspace{-0.05cm} 
F_{\mathcal{D}_+,\mathcal{D}_-}(\vx;\vz)+
\alpha_0 G_{\mathcal{D}_+,\mathcal{D}_-}(\vw;\vz)-\alpha_0^2\right]\\\label{eq:f1}
\hspace{-0.2cm}f_1(\vx):=&\hspace{-0.2cm}\max_{\alpha_1,\alpha_2\in\mathcal{I}} \hspace{-0.2cm} \mathbb{E}\hspace{-0.1cm} \left[\hspace{-0.2cm}
\begin{array}{ll}
F_{\mathcal{G}_1',\mathcal{G}_1}(\vx;\vz)+\alpha_1 G_{\mathcal{G}_1',\mathcal{G}_1}(\vw;\vz)-\alpha_1^2\\
+F_{\mathcal{G}_2,\mathcal{G}_2'}(\vx;\vz)+\alpha_2 G_{\mathcal{G}_2,\mathcal{G}_2'}(\vw;\vz)-\alpha_2^2
\end{array}
\hspace{-0.2cm}\right]\\\label{eq:f2}
\hspace{-0.2cm}f_2(\vx):=&\hspace{-0.2cm}\max_{\alpha_3,\alpha_4\in\mathcal{I}} \hspace{-0.2cm} \mathbb{E}\hspace{-0.1cm} \left[\hspace{-0.2cm}
\begin{array}{ll}
F_{\mathcal{G}_2',\mathcal{G}_2}(\vx;\vz)+\alpha_3 G_{\mathcal{G}_2',\mathcal{G}_2}(\vw;\vz)-\alpha_3^2\\
+F_{\mathcal{G}_1,\mathcal{G}_1'}(\vx;\vz)+\alpha_4 G_{\mathcal{G}_1,\mathcal{G}_1'}(\vw;\vz)-\alpha_4^2
\end{array}
\hspace{-0.2cm}\right]
\end{align}
\end{small}
and
\begin{small}
\begin{eqnarray*}
\label{eq:F0}
F_{\mathcal{D}_+,\mathcal{D}_-}(\vx;\vz)&=&F_{\mathcal{D}_+,\mathcal{D}_-}(\vw,a_0,b_0;\vz)\\
F_{\mathcal{G}_1',\mathcal{G}_1}(\vx;\vz)&=&F_{\mathcal{G}_1',\mathcal{G}_1}(\vw,a_1,b_1;\vz)\\
F_{\mathcal{G}_2,\mathcal{G}_2'}(\vx;\vz)&=&F_{\mathcal{G}_2,\mathcal{G}_2'}(\vw,a_2,b_2;\vz)\\
F_{\mathcal{G}_2',\mathcal{G}_2}(\vx;\vz)&=&F_{\mathcal{G}_2',\mathcal{G}_2}(\vw,a_3,b_3;\vz)\\
F_{\mathcal{G}_1,\mathcal{G}_1'}(\vx;\vz)&=&F_{\mathcal{G}_1,\mathcal{G}_1'}(\vw,a_4,b_4;\vz)
\end{eqnarray*}
\end{small}
with $F_{\mathcal{G},\mathcal{G}'}(\vw,a,b;\vz)$ and $G_{\mathcal{G},\mathcal{G}'}(\vw;\vz)$ defined in \eqref{eq:GFDef}.



\section{CONVEX CASE}
\label{sec:convexcase}
In this section, we introduce the stochastic feasible level-set (SFLS) method by \cite{lin2020data} for solving \eqref{eq:minmaxproblem} when the problem is convex. We make the following assumptions in this section.
\begin{assumption}\label{assum:convex}
$\mathbb{E}[F_{\mathcal{G},\mathcal{G}'}(\vx;\vz)]+\alpha\mathbb{E}[G_{\mathcal{G},\mathcal{G}'}(\vw;\vz)]$ is convex in $\vx$ for any sets $\mathcal{G}$ and $\mathcal{G}'$ and any $\alpha\in\mathbb{R}$.
\end{assumption}

This assumption holds when  $h_{\vw}(\vxi)=\vw^\top\vxi$.

\begin{assumption}\label{assum:deviation}
	There exists $\sigma>0$ such that 
 \small
	\begin{align}
	\E\left[\exp(|F_{\mathcal{G},\mathcal{G}'}(\vx;\vz)|^2/\sigma^2)\right]&\leq\exp(1),\\
	\E\left[\exp(|G_{\mathcal{G},\mathcal{G}'}(\vw;\vz)|^2/\sigma^2)\right]&\leq\exp(1),\\
	\E\left[\exp(\|\nabla F_{\mathcal{G},\mathcal{G}'}(\vx;\vz)\|_2^2/\sigma^2)\right]&\leq\exp(1),\\
	\E\left[\exp(\|\nabla G_{\mathcal{G},\mathcal{G}'}(\vw;\vz)\|_2^2/\sigma^2)\right]&\leq\exp(1)
	\end{align}
 \normalsize
for any sets $\mathcal{G}$ and $\mathcal{G}'$ and  any $\vx\in\mathcal{X}$, where $\nabla F_{\mathcal{G},\mathcal{G}'}(\vx;\vz)$ and $\nabla G_{\mathcal{G},\mathcal{G}'}(\vw;\vz)$ are the gradients of $F_{\mathcal{G},\mathcal{G}'}$ and $F_{\mathcal{G},\mathcal{G}'}$ with respect to $\vx$ and $\vw$, respectively. 
\end{assumption}

\begin{assumption}[Strict Feasibility] 
\label{assum1}
	There exists $\tilde \bx \in \mathcal{X}$ such that
	$\max\{f_1(\tilde \bx),f_2(\tilde \bx)\}<1+\kappa$. 
\end{assumption} 
As the following lemma shows, this assumption holds if $h_{\vw}(\cdot)$ becomes a constant mapping for some $\vw\in\mathcal{W}$. The proof is provided in Appendix~\ref{sec:lemma1}.
\begin{lemma}
\label{eq:feasible}
Assumption~\ref{assum1} holds if $c_1c_2^2\leq 0.5$ and there exists $\vw\in\mathcal{W}$ such that $h_{\vw}(\cdot)$ is a constant mapping.
\end{lemma}
The SFLS method relies on  the following \emph{level-set function}
\begin{align}
\label{eq:gcols}
H(r):=\min_{\bx\in\mathcal{X}}\Pc(r,\bx),
\end{align}
where $r\in\mathbb{R}$ is a \emph{level parameter} and 
\begin{align*}
\Pc(r,\bx)\hspace{-0.1cm}=\hspace{-0.1cm}\max\{f_0(\bx)-r,f_1(\bx)-1-\kappa,f_2(\bx)-1-\kappa\}.
\end{align*}

By lemmas 2.3.4 and 2.3.6 in \cite{nesterov2003introductory} and Lemma 1 in \cite{lin2018level}, $H(r)$ is non-increasing and convex and has an unique root at $r=f^*$. The SFLS method is essentially a root-finding procedure that generates a sequence of $r^{(k)}$, $k=0,1,\dots,$ approaching $f^*$ from the right. The update of $r^{(k)}$ requires the knowledge of $H(r)$ which is unknown. Typically, another algorithm is applied to \eqref{eq:gcols} to obtain an upper bound estimation of $H(r)$. This algorithm is called a stochastic oracle of $H(r)$ defined below.
\begin{definition}
	\label{def:feasOracle}
	Given $r>f^*$, $\epsilon_{\mathcal{A}}>0$, and $\delta\in(0,1)$, a \textbf{stochastic oracle} $\mathcal{A}(r, \epsilon_{\mathcal{A}}, \delta)$ returns $U(r)\in\mathbb{R}$ and $\bar \bx\in\mathcal{X}$ that satisfy the inequalities $\Pc(r,\bar \bx)- H(r)\leq\epsilon_{\mathcal{A}}$ and $|U(r)-H(r)| \leq \epsilon_{\mathcal{A}}$ with a probability of at least $1-\delta$.\looseness = -1
\end{definition}

Suppose a stochastic oracle  $\mathcal{A}$ exists, the SFLS method by~\cite{lin2020data} is presented in Algorithm~\ref{SFLS} with its convergence property given in Proposition~\ref{generalcomplexity}. 

\begin{algorithm}[t]
\caption{Stochastic Feasible Level-Set Method (SFLS)}\label{SFLS}
\begin{algorithmic}[1]
\State \textbf{Inputs:} A stochastic oracle $\mathcal{A}$, a level parameter $r^{(0)}>f^*$, an optimality tolerance $\epsilon_{\text{opt}}>0$, an oracle error $\epsoracle>0$, a probability $\delta\in(0,1)$, and a step length parameter $\theta>1$. 
\For{$k=0,1\cdots,$}
\State $\delta^{(k)}=\frac{\delta}{2^k}$
\State $(U(r^{(k)}),\bx^{(k)}) =\mathcal{A}(r^{(k)},\epsilon_{\mathcal{A}},\delta^{(k)})$
\If{$U(r^{(k)})\geq -\epsilon_{opt}$}
\State Halt and return $\bx^{(k)}$
\EndIf
\State $r^{(k+1)}\leftarrow r^{(k)}+U(r^{(k)})/(2\theta)$ and $k\leftarrow k+1$
\EndFor
\State \textbf{end for}
\end{algorithmic}
\end{algorithm}

\begin{proposition}[Theorem 5 in~\cite{lin2020data}]
\label{generalcomplexity}
Suppose $\epsilon_{\text{opt}}=-\frac{1}{\theta} H(r^{(0)})\epsilon$ and $\epsoracle=-\frac{\theta-1}{2\theta^2(\theta+1)}  H(r^{(0)}) \epsilon$ for $\epsilon\in(0,1)$. Algorithm~\ref{SFLS} generates a feasible solution at each iteration with a probability of at least $1-\delta$. Moreover, it returns an $\bx^{(k)}$ that is feasible and relative $\epsilon$-optimal, i.e., $(f_0(\bx^{(k)}) - f^*)/(f_0(\bx^{(0)}) - f^*)  \leq \epsilon$ with this probability in at most $\tilde O(\frac{1}{\epsilon^2})$ iterations. \footnote{Here and in the rest of the paper, $\tilde O$ suppresses the logarithmic factors in the order of magnitude.}
\end{proposition}


The remaining question is how to design an stochastic oracle $\mathcal{A}(r, \epsilon, \delta)$  satisfying Defintion~\ref{def:feasOracle}. Let $\tilde\by=(\tilde y_0,\tilde y_1,\tilde y_2)\in\Delta_3:=\{\tilde\by\in\mathbb{R}^3_+|\sum_{i=0}^2\tilde y_i=1\}$. With \eqref{eq:f0}, \eqref{eq:f1} and \eqref{eq:f2}, we can reformulate \eqref{eq:gcols} into
\begin{equation}
\label{eq:oldphi}
    H(r):=\min_{\bx\in\mathcal{X}}
\max_{\tilde\by\in\Delta_3, \valpha\in \mathcal{I}^5}\tilde{\phi}(\vx,\tilde{\vy},\valpha)
\end{equation}
where the definition of $\tilde{\phi}(\vx,\tilde{\vy},\valpha)$ is in Appendix~\ref{def_formula}.
This min-max optimization problem is not jointly concave in $\tilde{\vy}$ and $\valpha$ due to their product terms. As a result, the standard stochastic mirror descent method, e.g., \cite{nemirovski2009robust}, does not necessarily converge in theory if applied directly to \eqref{eq:oldphi}. Motivated by \cite{lin2018level_finitesum}, we equivalently convert this min-max problem above into a convex-concave min-max problem by changing variables. In particular, we define variables $\tilde\valpha=(\tilde\alpha_i)_{i=0}^4$ where
$
\tilde\alpha_0=\tilde y_0\alpha_0, \tilde\alpha_1=\tilde y_1\alpha_1, \tilde\alpha_2=\tilde y_1\alpha_2, \tilde\alpha_3=\tilde y_2\alpha_3, \tilde\alpha_4=\tilde y_2\alpha_4
$
and define $\by=(\tilde\by,\tilde\valpha)$ and
$$
\mathcal{Y}:=\left\{\by=(\tilde\by,\tilde\valpha)\Bigg|
\begin{array}{c}
\tilde \by\in\Delta_3, \tilde\alpha_0\in \tilde y_0\cdot\mathcal{I},\\
\tilde\alpha_1,\tilde\alpha_2\in \tilde y_1\cdot\mathcal{I},\tilde\alpha_3,\tilde\alpha_4\in\tilde  y_2\cdot\mathcal{I}
\end{array}
\right\}.
$$ 
Eliminating $\valpha$ by $\tilde\valpha$ in \eqref{eq:oldphi} gives
\begin{eqnarray}
\label{eq:minmax_old}
\min_{\bx\in\mathcal{X}}
\max_{\by\in\mathcal{Y}}\left\{
\phi(\bx,\by)-d_y(\by)
\right\},
\end{eqnarray}
where $\phi(\bx,\by):=\E[\Phi(\bx,\by,\vz)]$,
\begin{align}
\label{eq:dy}
d_y(\by):=&\frac{\tilde \alpha_0^2}{\tilde y_0}+\frac{\tilde\alpha_1^2}{\tilde y_1}+\frac{\tilde\alpha_2^2}{\tilde y_1}+\frac{\tilde\alpha_3^2}{\tilde y_2}+\frac{\tilde\alpha_4^2}{\tilde y_2},\\
\label{eq:Phi}
\Phi(\bx,\by,\vz):=&
\tilde\by^\top \vF(\bx,\vz)+\tilde\valpha^\top\vG(\vw,\vz),
\end{align}
\small
\begin{align*}
\vF(\bx,\vz)=\left(\begin{array}{c}
F_{\mathcal{D}_+,\mathcal{D}_-}(\vx;\vz)-r\\
F_{\mathcal{G}_1',\mathcal{G}_1}(\vx;\vz)+F_{\mathcal{G}_2,\mathcal{G}_2'}(\vx;\vz)-1-\kappa\\
F_{\mathcal{G}_2',\mathcal{G}_2}(\vx;\vz)+F_{\mathcal{G}_1,\mathcal{G}_1'}(\vx;\vz)-1-\kappa
\end{array}
\right)
\end{align*}
\normalsize
and
\small
\begin{align*}
\vG(\bw,\vz)=\left(\begin{array}{c}
G_{\mathcal{D}_+,\mathcal{D}_-}(\vw;\vz)\\
G_{\mathcal{G}_1,\mathcal{G}_1'}(\vw;\vz) \\ G_{\mathcal{G}_2',\mathcal{G}_2}(\vw;\vz)\\
G_{\mathcal{G}_2,\mathcal{G}_2'}(\vw;\vz) \\ G_{\mathcal{G}_1',\mathcal{G}_1}(\vw;\vz)
\end{array}
\right).
\end{align*}
\normalsize
We also slightly generalize \eqref{eq:minmax_new} to the following problem
\begin{eqnarray}
\label{eq:minmax_new}
\min_{\bx\in\mathcal{X}}
\max_{\by\in\mathcal{Y}}\left\{
\phi(\bx,\by)-d_y(\by)+d_x(\vx)
\right\},
\end{eqnarray}
where $d_x(\bx)=\frac{\hat\rho}{2}\|\vx-\tilde\vx\|_2^2$ for some $\hat\rho\geq0$ and some $\tilde\vx\in\mathcal{X}$. In this section, we focus on the convex case and only need to solve \eqref{eq:minmax_new} with $\hat\rho=0$. When we solve the weakly convex case later, we will set $\hat\rho>0$ and choose some  $\tilde\vx$.

Note that \eqref{eq:minmax_new} is a convex-concave min-max problem. In fact, except the term $d_y(\by)$, the objective function is linear in $\by$, which allows us to apply stochastic mirror descent (SMD) method. 
 The SMD method requires some distance generating function on $\mathcal{X}$ and $\mathcal{Y}$ and their corresponding \emph{Bregman divergences}. In our problem, the distance generating functions on $\mathcal{X}$ and $\mathcal{Y}$ are chosen as $\omega_x(\bx):=\frac{1}{2}\|\bx\|_2^2$ and $\omega_y(\by):=2(1+\sqrt{2}I)^2\left(\sum\limits_{i=0}^2\tilde y_i\ln{\tilde y_i}+\ln3\right)+d_y(\by)$
respectively, where $I:=\max_{\alpha\in\mathcal{I}}|\alpha|$. Function $\omega_y(\by)$ is specially designed for the set $\mathcal{Y}$ so, as we will show below, the iterates in the SMD method can be updated in closed-forms. Note that we can always choose $\mathcal{I}$ such that it satisfies \eqref{eq:intervalI} and is bounded. In fact, since $\mathcal{W}$ is compact and $\mathbb{E}[h_{\vw}(\bxi)|\mathcal{G}]$ is continuous in $\vw$, the intervals $\mathcal{I}_{\mathcal{D}_+, \mathcal{D}_-}$, $\mathcal{I}_{\mathcal{G}_1, \mathcal{G}_1'}$ and $\mathcal{I}_{\mathcal{G}_2,\mathcal{G}_2'}$ are all bounded, so we can also set $\mathcal{I}$ to be bounded. This ensures $I<+\infty$.

Let $\|\vx\|_x:=\|\vx\|_2$ and $\|\by\|_{y}:=\|\by\|_{1,2}:=\sqrt{\|\tilde\by\|_1^2+\|\tilde\valpha\|_2^2}$. It is clear that $\omega_x(\bx)$ is $1$-strongly convex on $\mathcal{X}$ with respect to $\|\vx\|_x$. It is shown by Lemma 2 in \cite{lin2018level_finitesum} that $\omega_y(\by)$ is $1$-strongly convex on $\mathcal{Y}$ with respect to $\|\by\|_{y}$. Hence, we can use them to define Bregman divergence
$V_x(\bx,\bx')=\omega_x(\bx)-[\omega_x(\bx')+\nabla\omega_x(\bx')^\top(\bx-\bx')]=\frac{1}{2}\|\bx-\bx'\|_2^2$ and 
\begin{align}\nonumber
	& V_y(\by,\by')
 :=\omega_y(\by)-[\omega_y(\by')+\nabla\omega_y(\by')^\top(\by-\by')]\\\label{eq:Distancey}
	=&2(1+\sqrt{2}I)^2\sum_{i=0}^2\tilde y_i\ln\left(\frac{\tilde y_i}{\tilde y_i'}\right) + \tilde y_0\left(\frac{\tilde \alpha_0}{\tilde y_0}-\frac{\tilde \alpha_0'}{\tilde y_0'}\right)^2\\
    & + \tilde y_1\sum^2_{i=1}\left(\frac{\tilde \alpha_i}{\tilde y_1}-\frac{\tilde \alpha_i'}{\tilde y_1'}\right)^2 + \tilde y_2\sum^4_{j=3}\left(\frac{\tilde \alpha_j}{\tilde y_2}-\frac{\tilde \alpha_j'}{\tilde y_2'}\right)^2 .\nonumber
\end{align}

With these Bregman divergences, we describe the SMD method in Algorithm~\ref{alg:smd}. The subproblems \eqref{eq:BregmanProx} and \eqref{eqn:compUBExpression} have closed-form solutions, which are characterized in Lemma \ref{lem:closedform} in Appendix~\ref{sec:closedform}.

\begin{algorithm}[t]
\caption{Stochastic Mirror Descent for \eqref{eq:minmax_new}} 
\label{alg:smd}
\begin{algorithmic}[1]
  \State {\bfseries Input:} Level parameter $r\in \mathbb{R}$, number of iterations $T$,  step size $\eta_t$ and $\tau_t$, $\hat\rho\geq0$ and  $\tilde\bx$. 
  \State Set $\vx^{(0)}=\mathbf{0}$, $\tilde\by^{(0)}=(\frac{1}{3},\frac{1}{3},\frac{1}{3})^\top$, $\tilde\valpha^{(0)}=\mathbf{0}$ and $\by^{(0)}=(\tilde\by^{(0)},\tilde\valpha^{(0)})$
  \For{$t=0$ {\bfseries to} $T-1$}
  \State Sample $\vz^{(t)}$.
  \State Compute stochastic gradients:
  \small
   \begin{align}
   \nonumber
 \vg_\vx^{(t)}=\nabla_x\Phi(\bx^{(t)},\by^{(t)},\vz^{(t)}),\ 
  \vg_\vy^{(t)}=\nabla_y\Phi(\bx^{(t)},\by^{(t)},\vz^{(t)})
 \end{align}
  \normalsize
 \State Primal-dual stochastic mirror descent:
  \normalsize
  \small
    \begin{align}
   \nonumber
    \vx^{(t+1)}=&\argmin_{\vx\in\mathcal{X}}\left \langle  \vg_\vx^{(t)},\vx \right \rangle+ \frac{\|\vx-\vx^{(t)}\|_2^2}{2\eta_t}+d_x(\vx)\\  \label{eq:BregmanProx}
     \vy^{(t+1)}=&\argmin_{\by\in\mathcal{Y}}-\left \langle  \vg_\vy^{(t)},\vy \right \rangle+\frac{V_y(\by,\by^{(t)})}{\tau_t}+d_y(\by) 
    \end{align}
  \normalsize
\State 	Compute a stochastic upper bound
\small
\begin{align}\label{eqn:compUBExpression}
&U(r):=\max_{\by\in\mathcal{Y}}\nonumber\\
&\left\{\frac{\sum_{t=0}^{T-1}\tau_t\left[\Phi(\bx^{(t)},\by,\vz^{(t)})-d_y(\by)+d_x(\vx^{(t)})\right]}{\sum_{t=0}^{T-1}\tau_t}\right\}.
		\end{align}
  \normalsize
  \EndFor
  \State {\bfseries Output:} $U(r)$ and $(\bar\vx ,\bar\vy)=\frac{1}{T}\sum_{t=0}^{T-1}(\vx^{(t)},\vy^{(t)})$.
\end{algorithmic}
\end{algorithm}

The convergence property of Algorithm~\ref{alg:smd} is well known (see, e.g., \cite{nemirovski2009robust,lin2020data}) and, in combination with Proposition~\ref{generalcomplexity},  it implies the total complexity of Algorithm~\ref{SFLS} as stated in the following theorem.
\begin{theorem}
\label{thm:convexcase}
\quad\\ Let $D_x:=\sqrt{\max_{\bx\in\mathcal{X}}\omega_x(\bx)-\min_{\bx\in\mathcal{X}}\omega_x(\bx)}$ and $D_y  := \sqrt{\max_{\by\in\mathcal{Y}}\omega_y(\by)-\min_{\by\in\mathcal{Y}}\omega_y(\by)}$. There exists a constant $M$ depending on $\sigma$, $\hat\rho$, $I$, $D_x$ and $D_y$ such that:
\begin{itemize}
\item Algorithm~\ref{alg:smd} is a stochastic oracle by Definition~\ref{def:feasOracle} if $T\geq \tilde O\left(\frac{1}{\epsilon_{\mathcal{A}}^2}\ln(\frac{1}{\delta})\right)$ and
\small
\begin{eqnarray}
\label{eq:smdstepsize}
\eta_t=2D_x^2/(M\sqrt{t+1}), \tau_t=2D_y^2/(M\sqrt{t+1}).
\end{eqnarray}
\normalsize
\item Suppose $\epsilon_{\text{opt}}$ and $\epsoracle$ are defined the same as in Proposition~\ref{generalcomplexity}. If Algorithm~\ref{alg:smd} is used as the stochastic oracle $\mathcal{A}$ with $\eta_t$ and $\tau_t$ defined as in \eqref{eq:smdstepsize} and $T\geq \tilde O\left(\frac{1}{\epsilon_{\mathcal{A}}^2}\ln(\frac{1}{\delta^{(k)}})\right)$ with $\delta^{(k)}$ defined in Algorithm~\ref{SFLS}. Algorithm~\ref{SFLS} returns a relative $\epsilon$-optimal and feasible solution with probability of at least $1-\delta$ after running at most $\tilde O\left(\frac{1}{\epsilon^2}\ln(\frac{1}{\delta})\right)$ stochastic mirror descent steps across all calls of $\mathcal{A}$.  
\end{itemize}
\end{theorem}
In Appendix~\ref{sec:mainproof}, we provide the definition of $M$ and the exact value of $T$ and we also give a brief discussion on how this theorem is obtained by applying the convergence results in~\cite{nemirovski2009robust,lin2020data} to Algorithm~\ref{alg:smd}. 

\section{WEAKLY-CONVEX CASE}
\label{sec:weaklyconvexcase}
In this section, we apply the proximal point techniques by \cite{boob2022stochastic,ma2020quadratically,jia2022first} to
extend the approach to the case where the objective and constraint functions in \eqref{eq:minmaxproblem} are weakly convex. 
\begin{definition}
\label{def:wcsc}
Given $h:\mathbb{R}\rightarrow\mathbb{R}\cup\{\infty\}$, we say $h$ is $\mu$-strongly convex for $\mu\geq0$ if 
\small
\begin{align*}
h(\vx)\geq h(\vx')+\vg^\top(\vx-\vx')+\frac{\mu}{2}\|\vx-\vx'\|_2^2
\end{align*}
\normalsize
for any $\vx,\vx'\in\mathcal{X}$ and any  $\vg\in\partial h(\bx)$, 
and we say $h$ is $\rho$-weakly convex for $\rho\geq0$ if 
\small
\begin{align*}
h(\vx)\geq h(\vx')+\vg^\top(\vx-\vx')-\frac{\rho}{2}\|\vx-\vx'\|_2^2
\end{align*}
\normalsize
for any $\vx,\vx'\in\mathcal{X}$ and any  $\vg\in\partial h(\bx)$. Here, $\partial h(\bx)$ is the subdifferential of $h$ at $\bx$.
\end{definition}
In this section, we do not assume Assumption~\ref{assum:convex} but assume Assumptions~\ref{assum:deviation} and \ref{assum1} and the following assumption.
\begin{assumption}\label{assum:wconvex}
	The following statements hold:
	\begin{itemize}
 \item[1.] $\mathbb{E}[F_{\mathcal{G},\mathcal{G}'}(\vx;z)]+\alpha\mathbb{E}[G_{\mathcal{G},\mathcal{G}'}(\vw;z)]$ is $\rho$-weakly convex in $\vx$ for any sets $\mathcal{G}$ and $\mathcal{G}'$ and any $\alpha\in\mathbb{R}$.
 \item[2.] There exist $\sigma_\epsilon>0$ and $\rho_\epsilon>0$ such that
 \small
 $$
 \min_{\bx'\in\mathcal{X}} \left\{\max_{i=1,2}f_i(\bx') -1-\kappa + \frac{\rho+\rho_\epsilon}{2}\|\bx' - \bx \|^2\right\}< -\sigma_\epsilon
 $$
 \normalsize
 for any $\bx\in\mathcal{X}$ with $\max_{i=1,2}f_i(\bx)-1-\kappa \leq \epsilon^2$. 
 \item[3.] $\|\vg\|\leq G$ for a constant $G$ for any $\vg\in\partial f_i(\bx)$ for $i=0,1,2$ and $\bx\in\mathcal{X}$.
 \end{itemize}
\end{assumption}

In Appendix~\ref{sec:assumption42}, we will provide a sufficient condition for Assumption~\ref{assum:wconvex}.2 to hold. In this case, the objective or the constraint functions can be non-convex, so finding an $\epsilon$-optimal solution is challenging in general. Hence, we target at finding a nearly $\epsilon$-stationary point defined below.

\begin{definition}
	\label{def:stationary}
A point $ \bx\in\mathcal{X}$ is called a \textbf{$\epsilon$-Karush-Kuhn-Tucker (KKT) point} of \eqref{eq:minmaxproblem} if there exist  Lagrangian multiplies $ \lambda_i\geq 0$ and $ \vg_i\in\partial f_i( \bx)$ for $i=1$ and $2$ such that
\begin{align*}
&\text{Dist}( \vg_0+ \lambda_1 \vg_1+ \lambda_2 \vg_2,-\mathcal{N}_{\mathcal{X}}( \bx))\leq \epsilon, \nonumber\\
&| \lambda_i(f_i( \bx)-1-\kappa)|\leq \epsilon^2,~ f_i( \bx)\leq 1+\kappa+ \epsilon, ~i=1,2,
\end{align*}
where $\mathcal{N}_{\mathcal{X}}(\bx)$ is the normal cone of $\mathcal{X}$ at $\bx$. Let $\hat{\rho}>\rho$. A point $\tilde\bx\in\mathcal{X}$ is called a \textbf{nearly $\epsilon$-stationary point} of \eqref{eq:minmaxproblem} if $\|\widehat\bx - \tilde\bx\|\leq\epsilon$ where 
	\begin{align}
	\label{eq:phix}
	\widehat\bx \equiv \argmin_{\bx'\in\mathcal{X}} &\ f_0(\bx') + \frac{\hat{\rho}}{2}\|\bx' - \tilde\bx\|_2^2,\\ \nonumber
	{s.t.}&\ f_i(\bx') + \frac{\hat{\rho}}{2}\|\bx' - \tilde\bx\|_2^2\leq 1+\kappa, i=1,2.
	\end{align}
\end{definition}
\begin{remark}
Since $\widehat\bx $  is optimal for \eqref{eq:phix},  there exist  Lagrangian multiplies $\widehat\lambda_i\geq 0$ and $ \widehat\vg_i\in\partial f_i( \widehat\bx)$ for $i=1$ and $2$ such that
\begin{align*}
&\text{Dist}( \widehat\vg_0+ \widehat\lambda_1 \widehat\vg_1+ \widehat\lambda_2 \widehat\vg_2,-\mathcal{N}_{\mathcal{X}}( \widehat\bx))\\
\leq&\hat{\rho}(1+\widehat\lambda_1+\widehat\lambda_2)\|\widehat\bx - \tilde\bx\|_2, \\
&| \widehat\lambda_i(f_i( \widehat\bx)-1-\kappa)|\leq\frac{\widehat\lambda_i\hat{\rho}}{2}\|\widehat\bx - \tilde\bx\|_2^2, \\
&f_i( \widehat\bx)\leq 1+\kappa, ~i=1,2.
\end{align*}
As discussed in \cite{jia2022first,boob2022stochastic,ma2020quadratically}, when $\widehat\lambda_i$ for $i=1$ and $2$ are bounded, a nearly $\epsilon$-stationary point $\tilde\bx$ is no more than $\epsilon$ away from $\widehat\bx $,  which is an $O(\epsilon)$-KKT point of \eqref{eq:minmaxproblem}. This justifies why  a nearly $\epsilon$-stationary point is a reasonable target for solving \eqref{eq:minmaxproblem} when the problem is non-convex. Different assumptions are considered in \cite{jia2022first,boob2022stochastic,ma2020quadratically} to ensure the boundness of $\widehat\lambda_i$. This paper follows \cite{ma2020quadratically} by assuming Assumption~\ref{assum:wconvex}.2 and the boundness of $\widehat\lambda_i$ under this assumption follows Lemma 1 in \cite{ma2020quadratically}.
\end{remark}


Next we apply the inexact quadratically regularized constrained (IQRC) method by
\cite{ma2020quadratically} to \eqref{eq:minmaxproblem}, which is given in Algorithm~\ref{alg:iqrc}. This algorithm requires an oracle define below.

\begin{definition}
		\label{def:subroutine}
  Given $\tilde\vx\in\mathcal{X}$, $\hat\rho>0$, $\hat\epsilon>0$, $\delta\in(0,1)$,  a \textbf{stochastic oracle} $\mathcal{B}(\tilde\bx,\hat\rho,\hat\epsilon,\delta)$ returns  $\bx'\in\mathcal{X}$ such that, with a probability of at least $1-\delta$, $\bx'$ is an $\hat\epsilon^2$-feasible and $\hat\epsilon^2$-optimal solution of \eqref{eq:phix}.
	\end{definition}

\begin{algorithm}[tb]
		\caption{Inexact Quadratically Regularized Constrained Method}
		\label{alg:iqrc}
		\begin{algorithmic}[1]
			\State   {\bfseries Input:} An $\epsilon^2$-feasible solution $\tilde\bx^{(0)}$, $\rho+\rho_\epsilon\geq\hat\rho>\rho$, $\delta\in(0,1)$, $\hat\epsilon=\min\Big\{1,\sqrt{\frac{\hat\rho-\rho}{4}}\Big(\frac{G+2\hat\rho D_x }{\sqrt{2 \sigma_{\epsilon} (\hat{\rho} - \rho)}}+1\Big)^{-\frac{1}{2}}\Big\}\epsilon $, and the number of iterations $S$.
			\For{$s=0,\dots,S-1$}
			\State  Compute $\tilde\bx^{(s+1)}=\mathcal{B}(\tilde\bx^{(s)},\hat\rho,\hat\epsilon,\frac{\delta}{S})$
	  \EndFor
			\State  {\bfseries Output:} $\tilde\bx^{(R)}$ where $R$ is a random index uniformly sampled from $\lbrace 0,\dots,S \rbrace$.
		\end{algorithmic}
	\end{algorithm}

 \begin{figure*}
\begin{tabular}[h]{@{}c|ccc@{}}
		& Group-AUC Fairness & Inter-Group Pairwise Fairness & Intra-Group Pairwise Fairness\\
		\hline \vspace*{-0.1in}\\		
		\raisebox{7ex}{\small{\rotatebox[origin=c]{90}{a9a}}}
		& \hspace*{-0.01in}\includegraphics[width=0.33\textwidth]{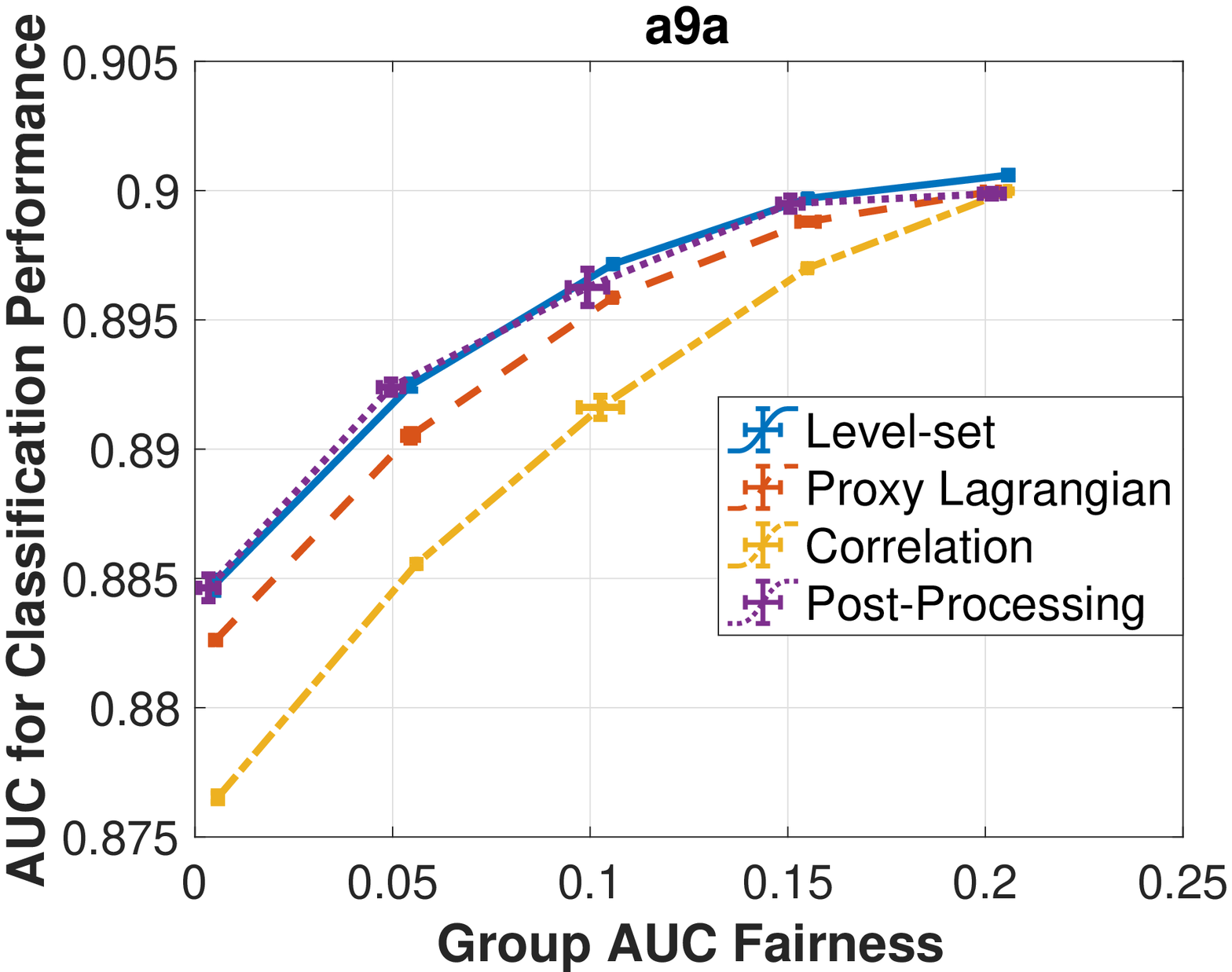}\vspace*{-0.01in}
		& \hspace*{-0.2in}\includegraphics[width=0.33\textwidth]{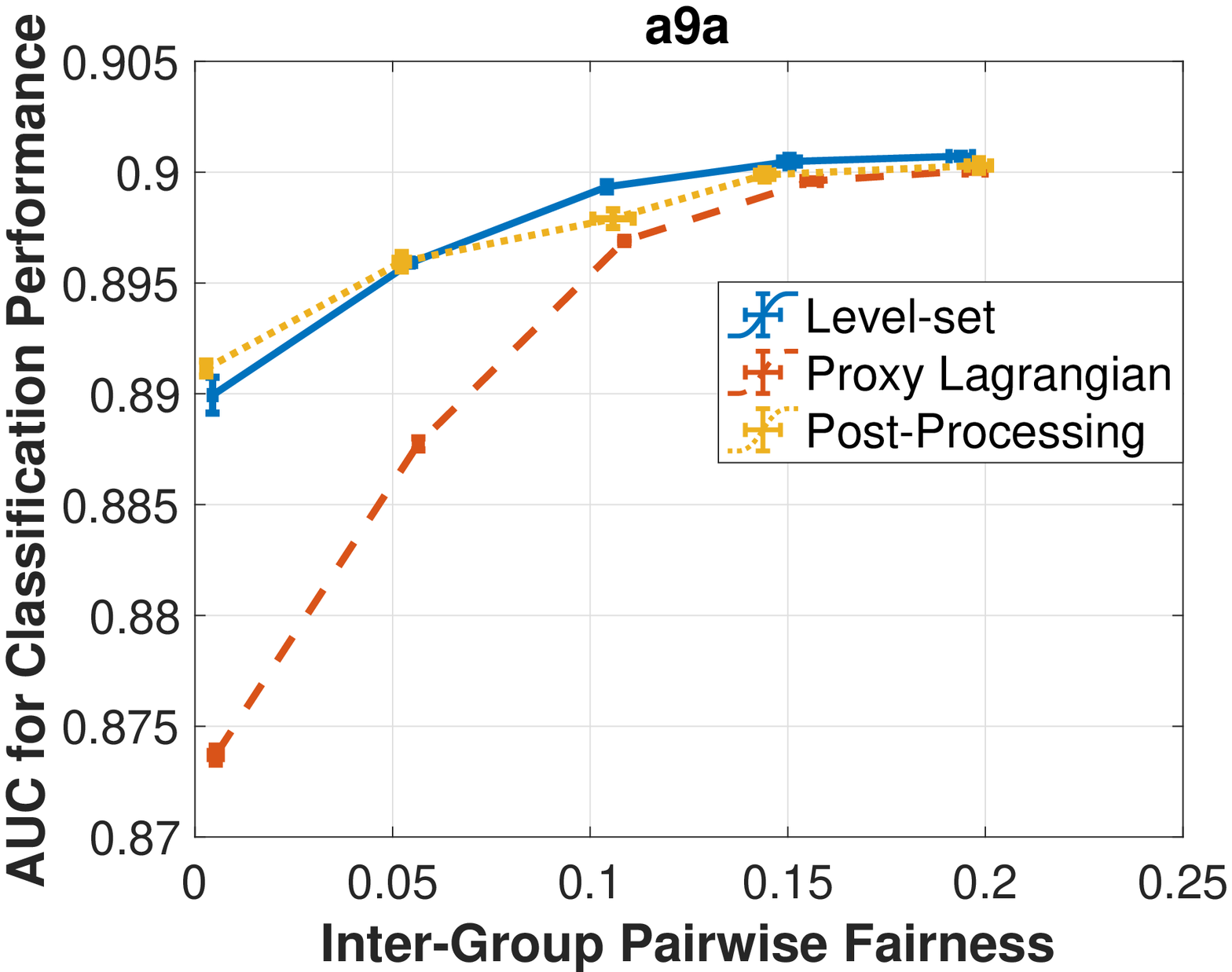}\vspace*{-0.01in}
		& \hspace*{-0.2in}\includegraphics[width=0.33\textwidth]{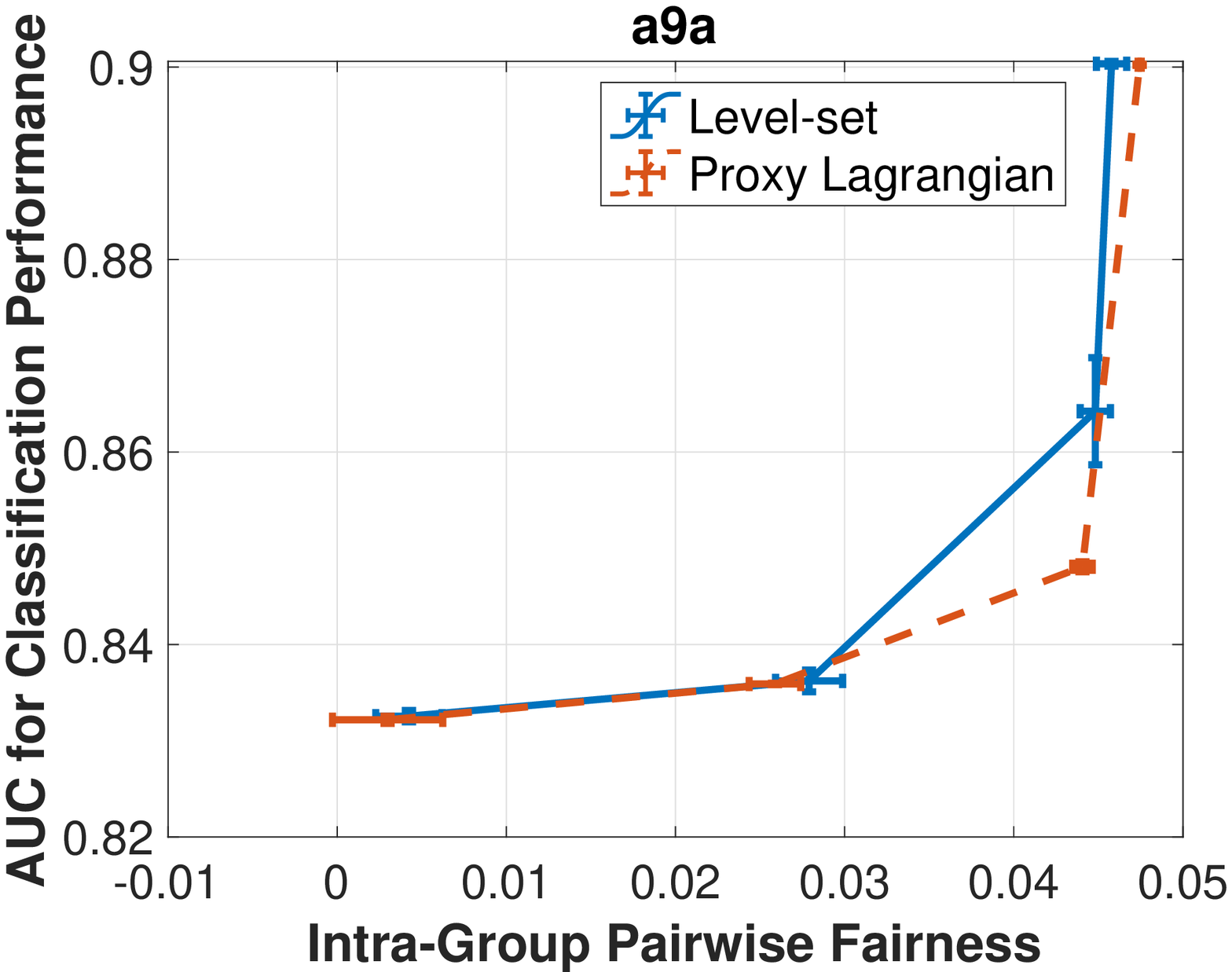}\vspace*{-0.01in}\\
		
		\raisebox{7ex}{\small{\rotatebox[origin=c]{90}{bank}}}
		& \hspace*{-0.01in}\includegraphics[width=0.33\textwidth]{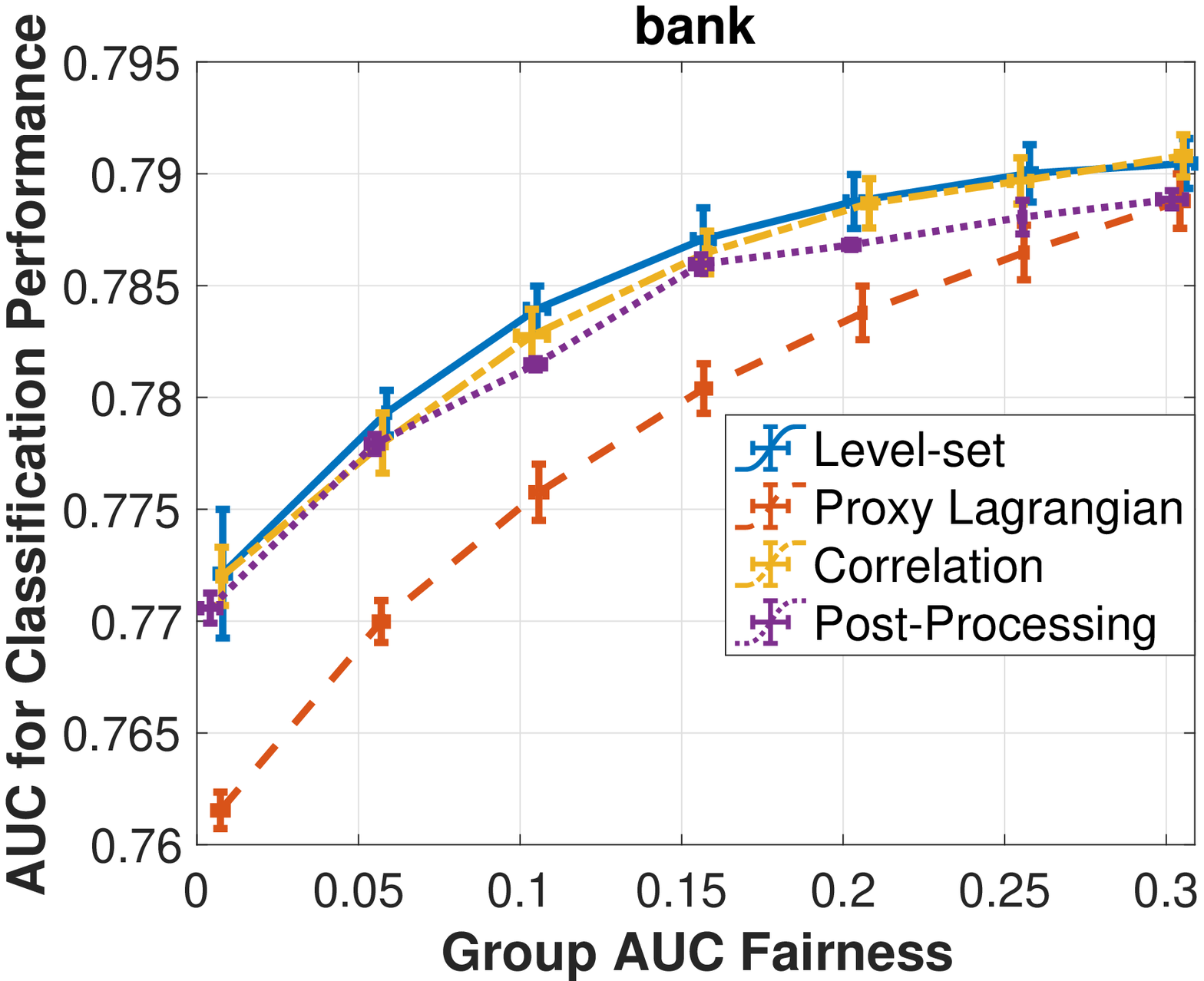}\vspace*{-0.01in}
		& \hspace*{-0.2in}\includegraphics[width=0.33\textwidth]{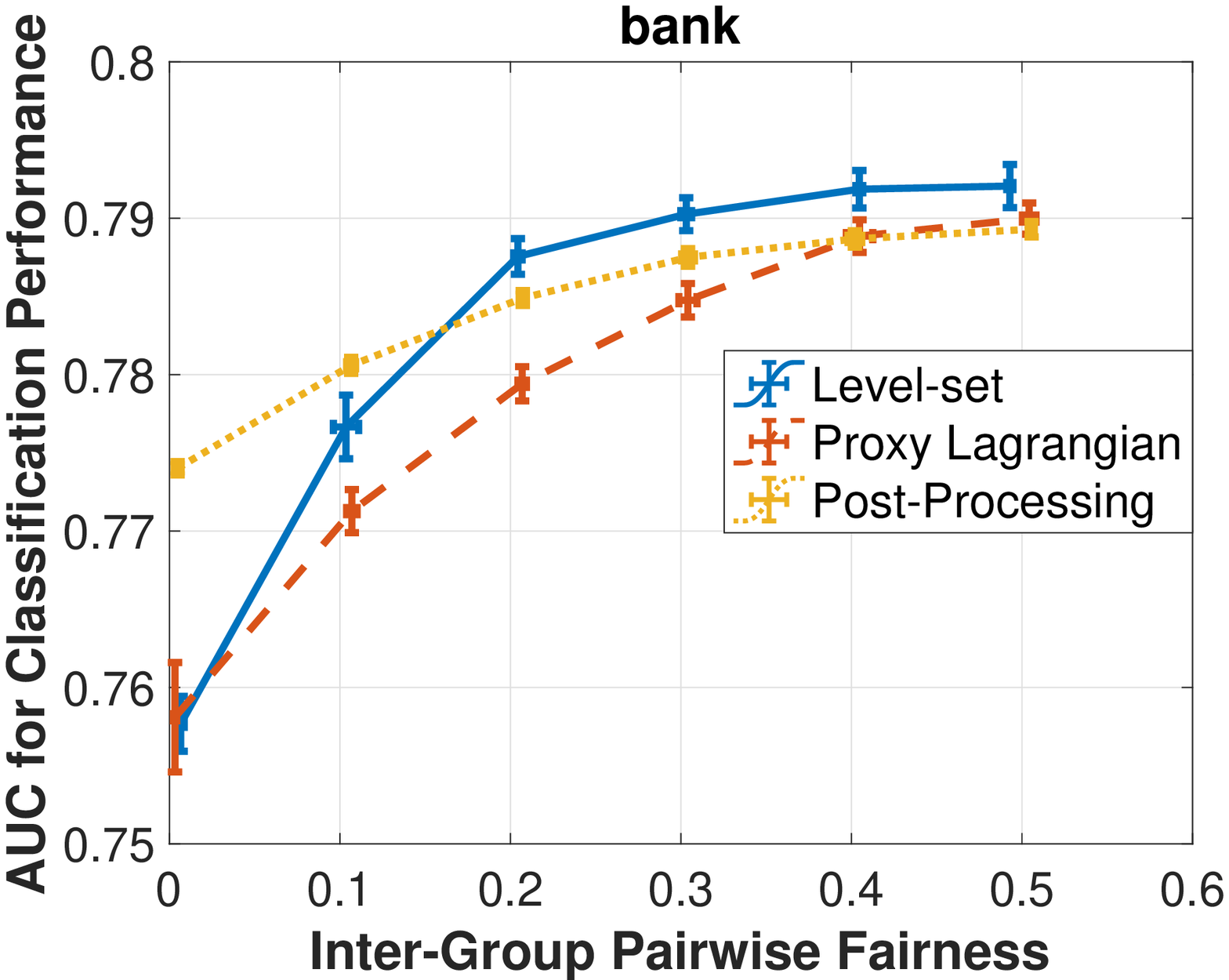}\vspace*{-0.01in}
		& \hspace*{-0.2in}\includegraphics[width=0.33\textwidth]{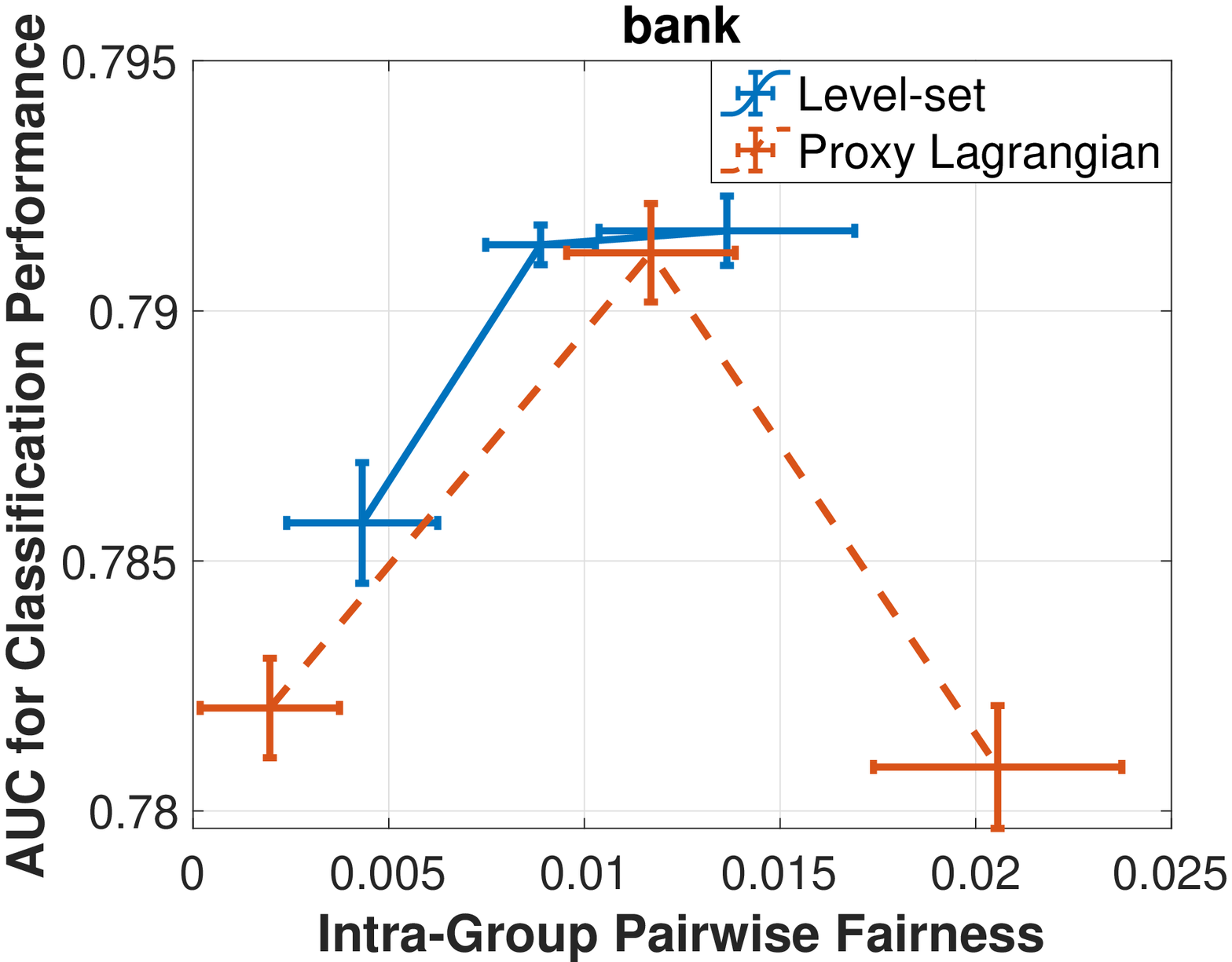}\vspace*{-0.01in}\\
		
  \end{tabular}
	\caption{Pareto frontiers by each method on testing set in convex case (see Appendix \ref{sec:compas} for COMPAS dataset).}
	\label{fig:figure}
\end{figure*}

According to the definition of this oracle, in its iteration $t$, Algorithm~\ref{alg:iqrc} needs to find $\hat\epsilon^2$-feasible and $\hat\epsilon^2$-optimal solution of subproblem \eqref{eq:phix} with $\tilde\bx=\tilde\bx^{(s)}$. Since \eqref{eq:phix} is convex when $\hat\rho>\rho$, Algorithm~\ref{SFLS} can be used as an oracle $\mathcal{B}$. To do so, we need to derive and solve the level-set subproblem \eqref{eq:gcols} corresponding to \eqref{eq:phix}, which is
\begin{align}
	\label{eq:gcolshat}
	\widehat H(r):=\min_{\bx\in\mathcal{X}}\Big\{\Pc(r,\bx)+\frac{\hat\rho}{2}\|\vx-\tilde\vx\|_2^2\Big\}.
\end{align}
Following the same step as in Section~\ref{sec:convexcase}, \eqref{eq:gcolshat} can be reformulated as \eqref{eq:minmax_new}. Recall that we set $\hat\rho=0$ in Section~\ref{sec:convexcase} when the problem is convex, but here we set $\hat\rho>\rho$ because of non-convexity.

According to Theorem~\ref{thm:convexcase}, when Algorithm~\ref{alg:smd} is used as the oracle $\mathcal{A}$ in Algorithm~\ref{SFLS}, Algorithm~\ref{SFLS} becomes an oracle $\mathcal{B}$ for Algorithm~\ref{alg:iqrc} with an iteration complexity of $\tilde O(\frac{1}{\hat\epsilon^4})$=$\tilde O(\frac{1}{\epsilon^4})$. According to Theorem 1 in \cite{ma2020quadratically}, Algorithm~\ref{alg:iqrc} finds a nearly $\epsilon$-stationary point of \eqref{eq:minmaxproblem} in $O(\frac{1}{\epsilon^2})$ iterations with $\mathcal{B}$ called once in each iteration. Combining these two results, we know that the total iteration complexity of Algorithm~\ref{alg:iqrc} is $\tilde O(\frac{1}{\epsilon^4})\times O(\frac{1}{\epsilon^2})= \tilde O(\frac{1}{\epsilon^6})$. This is formally stated in the following theorem. The proof is omitted since this theorem can be easily obtained from the existing results according to the discussion above. 
\begin{theorem}
		\label{thm:Yu}
Suppose Algorithm~\ref{alg:iqrc} uses Algorithm~\ref{SFLS} as oracle $\mathcal{B}$ and $\epsilon_{\text{opt}}$ and $\epsoracle$ in Algorithm~\ref{SFLS} are set as in Proposition~\ref{generalcomplexity} except that $H$ is replaced by $\widehat H$ in \eqref{eq:gcolshat}. Also, suppose Algorithm~\ref{SFLS} uses Algorithm~\ref{alg:smd} as oracle $\mathcal{A}$ and $\eta_t$, $\tau_t$ and $T$ are set as in Theorem~\ref{thm:convexcase}.  Algorithm~\ref{alg:iqrc} returns $\tilde\bx^{(R)}$ as a nearly $\epsilon$-stationary point of~\eqref{eq:minmaxproblem} within $\tilde O(\frac{1}{\epsilon^6})$ stochastic mirror descent steps across all calls of $\mathcal{B}$.
\end{theorem}

\section{NUMERICAL EXPERIMENTS}
\label{sec:exp}
In this section, we demonstrate the effectiveness of the proposed approaches for AUC optimization subject to the AUC-based fairness constraints given in Examples \ref{ex:GroupAUC}, \ref{ex:xAUC} and \ref{ex:intraAUC} in Section~\ref{sec:preliminary}. All experiments are conducted on a computer with the CPU 2GHz Quad-Core Intel Core i5 and the GPU NVIDIA GeForce RTX 2080 Ti. 

\textbf{Datasets Information.} The experiments are conducted using three public datasets: \textit{a9a}~\citep{kohavi1996,Dua:2019,chang2011libsvm}, \textit{bank}~\citep{moro2004,Dua:2019,chang2011libsvm} and \textit{COMPAS}~\citep{angwin2016compas,fabris2022}. Details about these datasets can be found in Appendix~\ref{sec:data}.

\textbf{Baselines.} We compare our methods with three baselines, the proxy-Lagrangian method~\citep{cotter2019optimization}, the correlation-penalty method~\citep{beutel2019fairness} and the post-processing method~\citep{kallus2019fairness}. The description of each baseline is provided in Appendix~\ref{sec:baseline}.

\begin{figure*}
\begin{tabular}[h]{@{}c|ccc@{}}
		& Group-AUC Fairness & Inter-Group Pairwise Fairness & Intra-Group Pairwise Fairness\\
		\hline \vspace*{-0.1in}\\		
		\raisebox{7ex}{\small{\rotatebox[origin=c]{90}{a9a}}}
		& \hspace*{-0.01in}\includegraphics[width=0.33\textwidth]{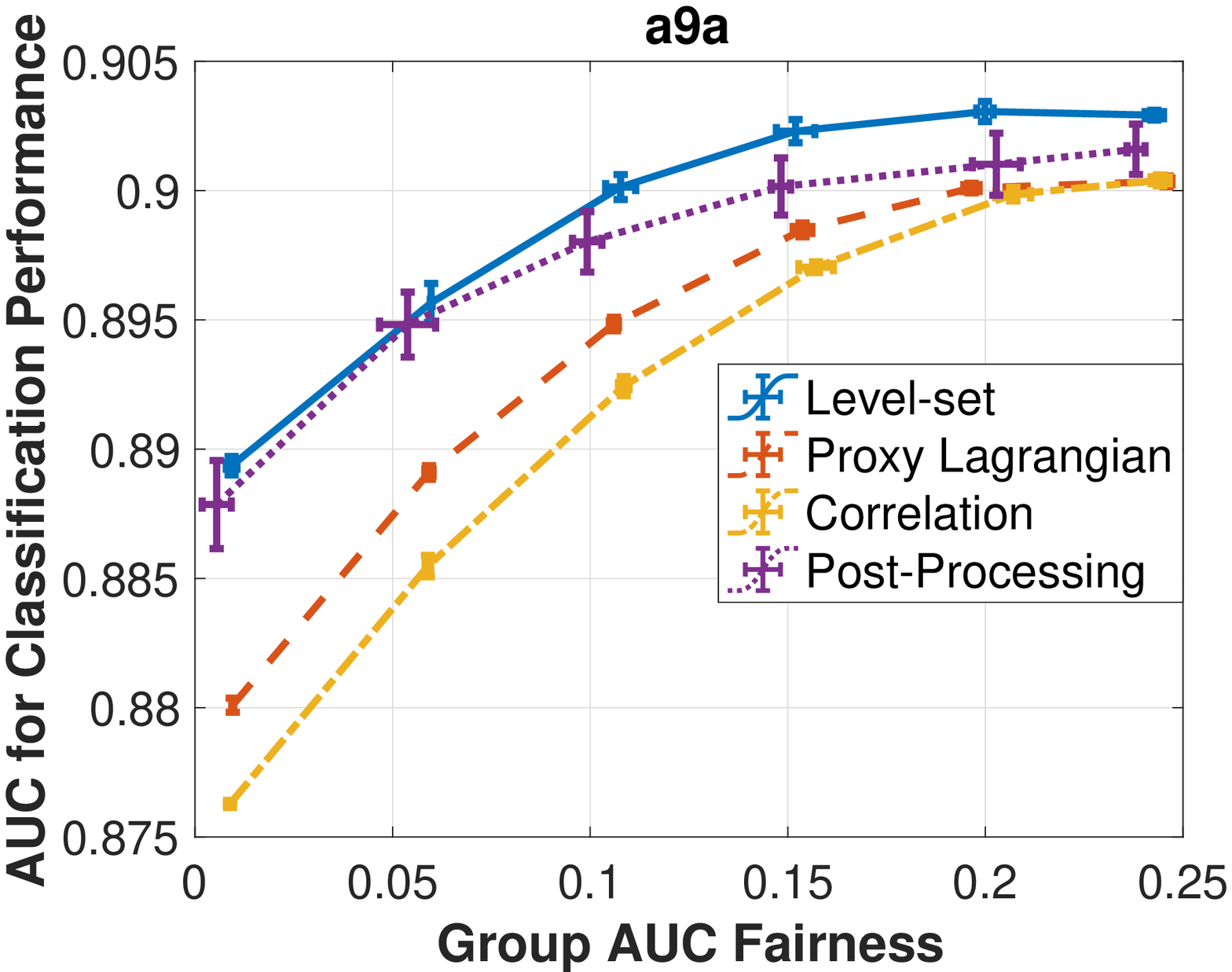}\vspace*{-0.01in}
		& \hspace*{-0.2in}\includegraphics[width=0.33\textwidth]{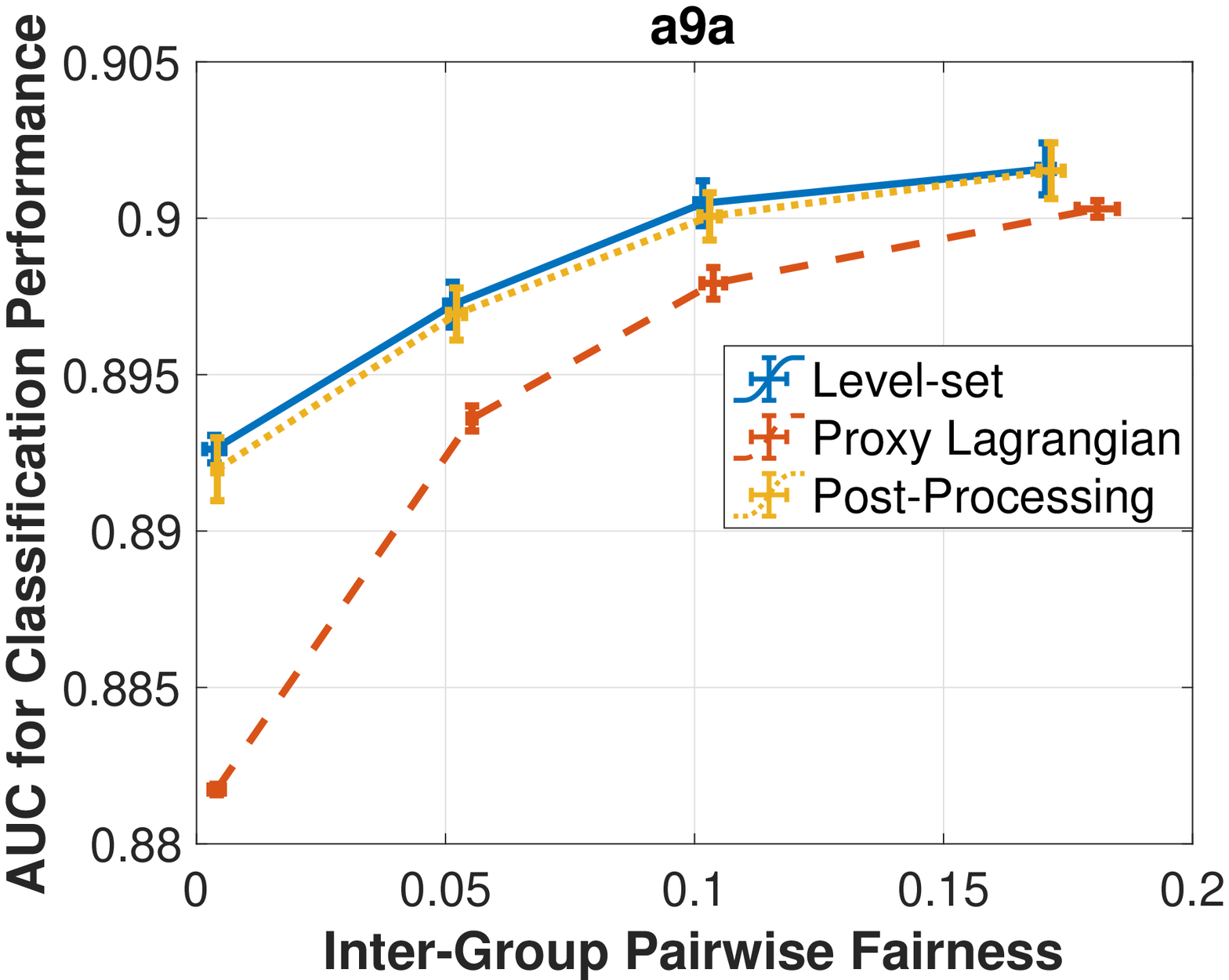}\vspace*{-0.01in}
		& \hspace*{-0.2in}\includegraphics[width=0.33\textwidth]{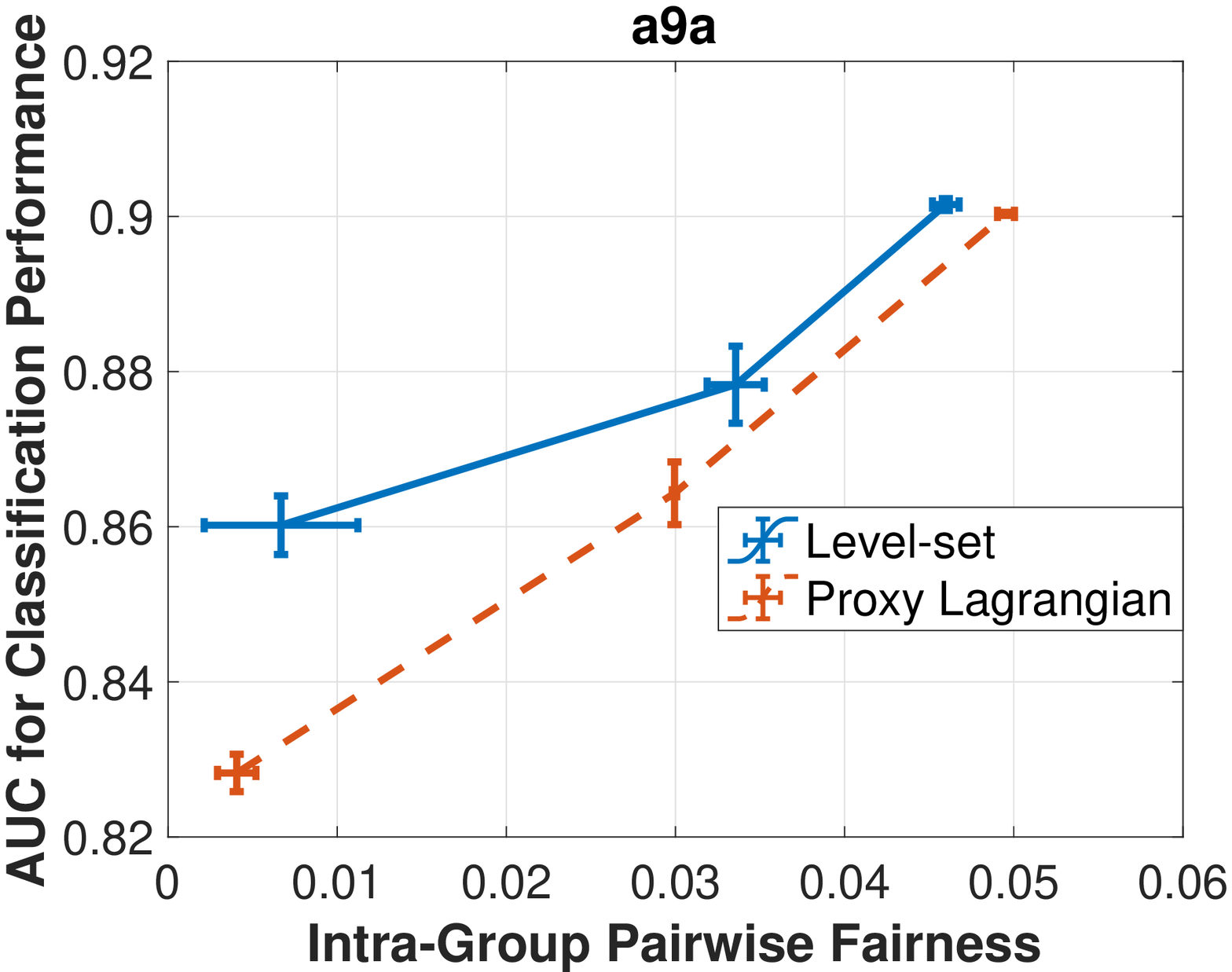}\vspace*{-0.01in}\\
		
		\raisebox{7ex}{\small{\rotatebox[origin=c]{90}{bank}}}
		& \hspace*{-0.01in}\includegraphics[width=0.33\textwidth]{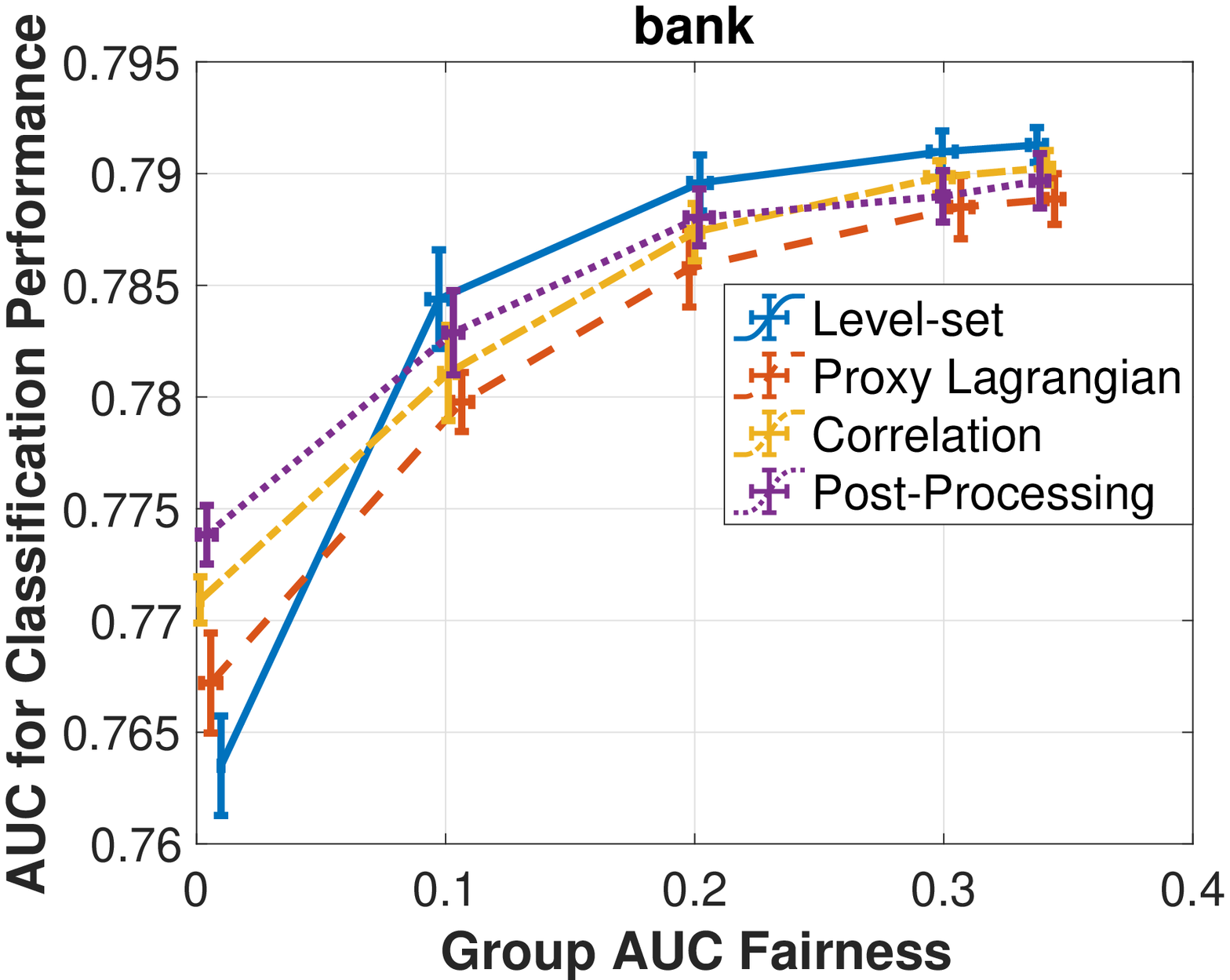}\vspace*{-0.01in}
		& \hspace*{-0.2in}\includegraphics[width=0.33\textwidth]{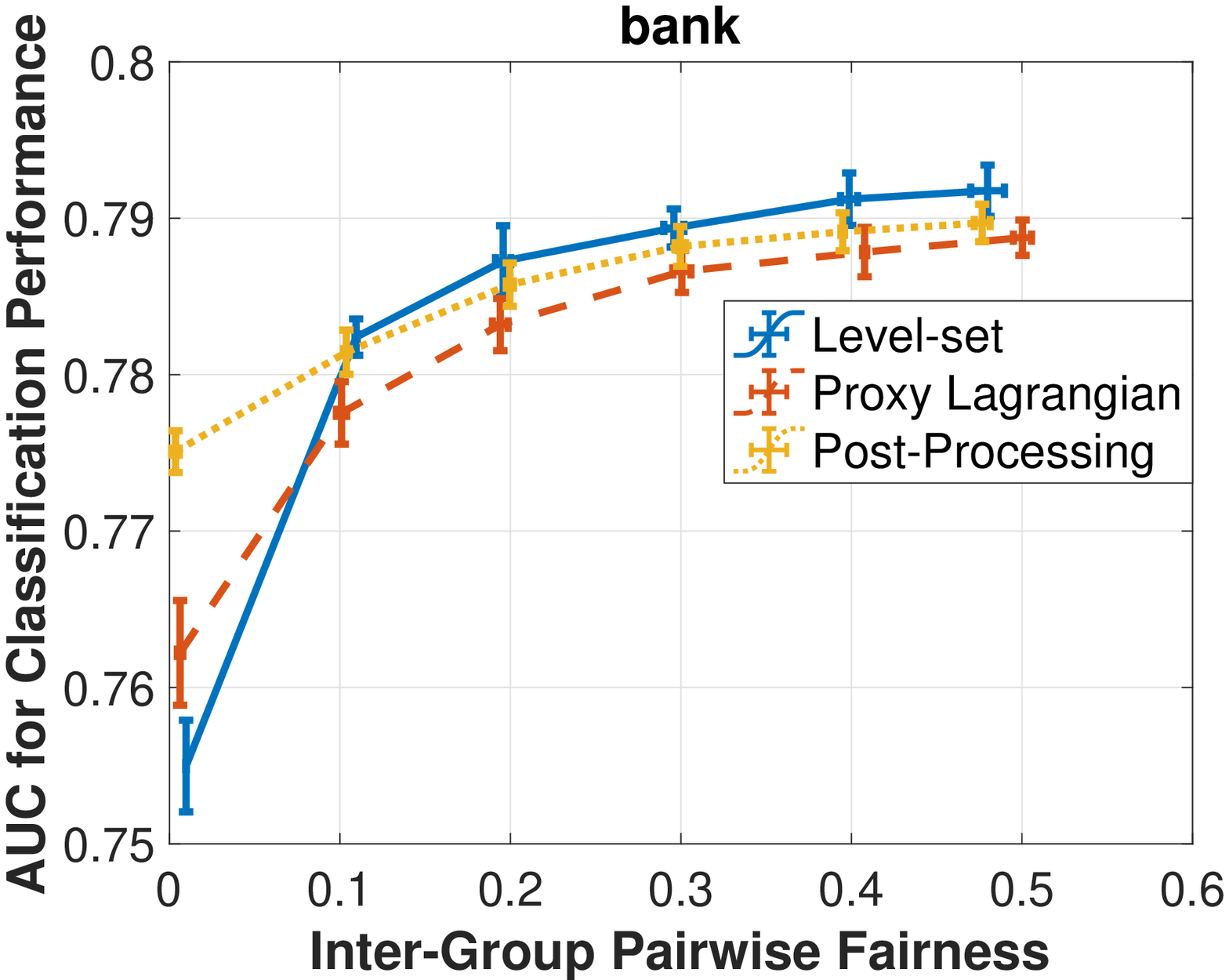}\vspace*{-0.01in}
		& \hspace*{-0.2in}\includegraphics[width=0.33\textwidth]{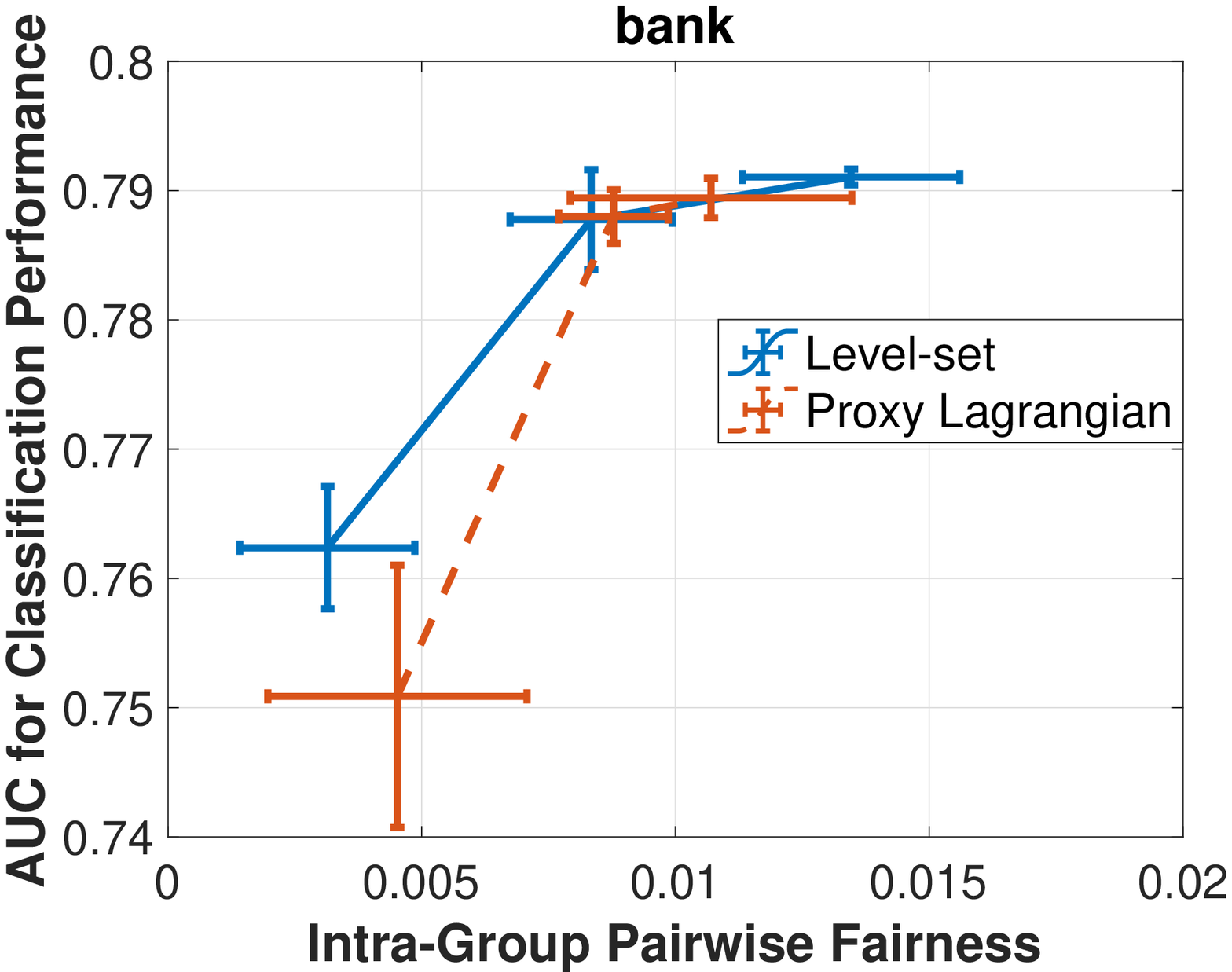}\vspace*{-0.01in}\\
		
  \end{tabular}
	\caption{Pareto frontiers by each method on testing set in weakly-convex case (see Appendix \ref{sec:compas} for COMPAS dataset).}
	\label{fig:figure_noncon}
	\vspace{-0.1in}
\end{figure*}

\textbf{Convex case.} For convex case, we consider a linear model, i.e., $h_{\vw}(\vxi)=\vxi^\top \vw$. A smaller $\kappa$ in \eqref{eqn:problem} makes the model more fair in terms of the corresponding fairness metric but may compromise the classification performance in terms of AUC. Hence, we varies $\kappa$ in \eqref{eqn:problem} so each method in comparison will generate a Pareto frontier, showing the trade-off between performance and fairness. 

For the three baselines and our algorithm, the process to tune the hyper-parameters is explained in
Appendix~\ref{sec:tune}. We then evaluate AUC and the fairness metric of the output model on testing set and report the Pareto frontiers by each method in Figure \ref{fig:figure}. We repeat each experiment five times with different random seeds and report the standard errors of the AUC scores and the fairness metrics through the error bars on each curve. Due to the page limit, we postpone the plots of \textit{COMPAS} dataset to Appendix \ref{sec:compas}.

\textbf{Weakly-convex case.} For weakly-convex case, we choose $h_{\vw}$ to be a two-layer neural network with 10 hidden neurons and
the sigmoid activation functions. The process of tuning hyperparameters is in Appendix~\ref{sec:tune}. 
In the non-convex case, the original proxy-Lagrangian method in~\cite{cotter2019optimization} updates $\vw$ through an  approximate  Bayesian optimization oracle, which can solve a non-convex problem with a reasonably small optimality gap. Here, we directly perform one stochastic gradient descent step to update $\vw$ just as in the convex case because it is unclear how to design such an oracle due to non-convexity. The Pareto frontiers in weakly-convex case are reported with error bars in Figure \ref{fig:figure_noncon}. 

It can be observed from Figures~\ref{fig:figure} and \ref{fig:figure_noncon} that the level-set method performs better than the other three baselines  when $\kappa$ is not too small. When $\kappa$ is small, the level-set method is less efficient in trading performance for fairness on the \textit{bank} dataset. This is likely because the approximation gap between \eqref{eq:minmaxproblem} and \eqref{eqn:problem} is large on this dataset. As a result, we have to use a very small $\kappa$ in \eqref{eq:minmaxproblem} in order to achieve the targeted fairness level in the original problem \eqref{eqn:problem}, which leads to very restrictive constraints in \eqref{eq:minmaxproblem} and harms the classification performance in terms of AUC.

\section{CONCLUSION and LIMITATION}
\vspace{-0.1in}
We consider AUC optimization subject to a class of AUC-based fairness constraints, which includes most of the existing threshold-agnostic and comparison-based fairness metrics in literature. When solving this problem in an online setting where the data arrives sequentially, the existing optimization methods need to receive at least a pair of data points to update the model, which may not be allowed by the order of data's arrivals. In addition, when the original problem is formulated using an empirical distribution in an off-line setting, the computational cost becomes quadratic in data size due to the definition of AUC. This computational cost is too high when the data is large. 

To address these computational challenges, we reformulated this problem into a min-max optimization problem subject to min-max constraints using a quadratic loss function to approximate the AUCs in the objective and constraint functions. The new optimization formulation allows the model to be updated in an online fashion with one data point arriving each time. In the off-line setting, the new formulation also reduces the computational cost to only linear in data size. By introducing a novel Bregman divergence after changing variables, we show that existing stochastic optimization algorithms can be applied to the new formulation in the convex and weakly convex cases. In the numerical experiments, we observe an efficient trade off between classification performance and fairness by the models created by our approaches. 

However, we acknowledge that our formulation only works for the quadratic loss function. It is our future work to further extend our methods for a general loss function.  








\subsubsection*{Acknowledgements}
This work was jointly supported by the University of Iowa Jumpstarting Tomorrow Program and NSF award 2147253.

\bibliographystyle{abbrvnat}
\bibliography{ref}




\appendix
\onecolumn

\section{EXAMPLES of FAIRNESS METRICS SATISFYING DEFINITION~\ref{def:AUCmetric}}
\label{sec:example}
In this section, we present five examples of fairness metrics that satisfy Definition~\ref{def:AUCmetric} and thus can be applied as fairness constraints in \eqref{eqn:problem} and solved by the optimization algorithms in this paper. In the discussion below, we assume all data points are ranked decreasingly in $h_{\vw}(\vxi)$ so  $h_{\vw}(\vxi)> h_{\vw}(\vxi')$ means $z$ is ranked higher than $z'$.

\begin{example}[Group AUC Fairness]
\label{ex:GroupAUC}
Let $\mathcal{G}_1=\{\vz|\gamma=1\}$, $\mathcal{G}_1'=\{\vz|\gamma=-1\}$ and $\mathcal{G}_2=\mathcal{G}_2'=\mathbb{R}^{p+2}$ (so $\text{AUC}_\vw(\mathcal{G}_2,\mathcal{G}_2')=0.5$). The AUC-based fairness metric becomes  
$
|\text{Pr}(h_{\vw}(\vxi)>h_{\vw}(\vxi')|\gamma=1,\gamma'=-1)-0.5|.
$
When it is small, a random data point from the protected group is ranked above a random data point from the unprotected group with nearly $50\%$ probability. In other words, if we use $h_{\vw}(\vxi)$ to predict sensitive variable $\gamma$, it must has a poor prediction performance in terms of AUC w.r.t. $\gamma$ (instead of $\zeta$).
\end{example}

\begin{example}[Inter-Group Pairwise Fairness]
\label{ex:xAUC}
Let $\mathcal{G}_1=\{\vz|\zeta=1,\gamma=1\}$, $\mathcal{G}_1'=\{\vz|\zeta=-1,\gamma=-1\}$,  $\mathcal{G}_2=\{\vz|\zeta=1,\gamma=-1\}$ and $\mathcal{G}_2'=\{\vz|\zeta=-1,\gamma=1\}$. In this case, the AUC-based fairness metric becomes the cross-AUC in \cite{kallus2019fairness}, which is also called inter-group pairwise fairness~\citep{beutel2019fairness}. When it is small, the probability of a random positive data point being ranked above a random negative data point from the opposite group is nearly independent of the group. 
\end{example}

\begin{example}[Intra-Group Pairwise Fairness]
\label{ex:intraAUC}
Let $\mathcal{G}_1=\{\vz|\zeta=1,\gamma=1\}$, $\mathcal{G}_1'=\{\vz|\zeta=-1,\gamma=1\}$,  $\mathcal{G}_2=\{\vz|\zeta=1,\gamma=-1\}$ and $\mathcal{G}_2'=\{\vz|\zeta=-1,\gamma=-1\}$. In this case, the AUC-based fairness metric becomes the intra-group pairwise fairness introduced by~\cite{beutel2019fairness}. When it is small, the probability of a random positive data point being ranked above a random negative data point from the same group is nearly independent of the group. In other words, the classical AUCs (w.r.t. class labels) evaluated separately on each  group are similar. 
\end{example}

\begin{example}[Average Equality Gaps]
	\label{ex:AEG}
	Let $\mathcal{G}_1=\{\vz|\zeta=1,\gamma=1\}$, $\mathcal{G}_1'=\{\vz|\zeta=1\}$ and $\mathcal{G}_2=\mathcal{G}_2'=\mathbb{R}^{p+2}$. The AUC-based fairness metric becomes the positive average equality gap introduced by~\cite{borkan2019nuanced}, i.e., 
	$
	|\text{Pr}(h_{\vw}(\vxi)>h_{\vw}(\vxi')|\gamma=1,\zeta=1,\zeta'=1)-0.5|.
	$
	Similar to Example 1, when this value is small, a random positive data point from the protected group is ranked above a random positive data from the whole dataset with nearly 50\% probability.  Similarly, the negative average equality gap by~\cite{borkan2019nuanced} is obtained when $\mathcal{G}_1=\{\vz|\zeta=-1,\gamma=1\}$, $\mathcal{G}_1'=\{\vz|\zeta=-1\}$ and $\mathcal{G}_2=\mathcal{G}_2'=\mathbb{R}^{p+2}$. 
 In this case,  the AUC-based fairness metric becomes
	$
	|\text{Pr}(h_{\vw}(\vxi)>h_{\vw}(\vxi')|\gamma=1,\zeta=-1,\zeta'=-1)-0.5|.
	$
	It has the similar interpretation as the positive average equality gap. 
\end{example}

\begin{example}[BPSN AUC and BNSP AUC]
	\label{ex:BPSNBNSP}
	When $\mathcal{G}_1=\{\vz|\zeta=1\}$ and $\mathcal{G}_1'=\{\vz|\zeta=-1,\gamma=1\}$, $\text{AUC}_\vw(\mathcal{G}_1,\mathcal{G}_1')$ becomes the background positive subgroup negative (BPSN) AUC in \cite{borkan2019nuanced}.  	When $\mathcal{G}_2=\{\vz|\zeta=1,\gamma=1\}$ and $\mathcal{G}_2'=\{\vz|\zeta=-1\}$, $\text{AUC}_\vw(\mathcal{G}_2,\mathcal{G}_2')$ becomes the background negative subgroup positive (BNSP) AUC in \cite{borkan2019nuanced}. One fairness metric introduced by \citet{borkan2019nuanced} is the absolute difference between the BPSN AUC and the BNSP AUC, which is exactly \eqref{eq:AUCmetric} w.r.t $\mathcal{G}_1$, $\mathcal{G}_1'$, $\mathcal{G}_2$ and $\mathcal{G}_2'$ chosen above. When this metric is small, the probability of a random positive data point from the whole dataset being ranked above a random negative data point from the protected group is close to the probability of a random positive data point from the protected group being ranked above a random negative data point from the whole dataset.
\end{example}

\section{TECHNICAL LEMMAS AND THEIR PROOFS}
\label{sec:lemma1}
In this section, we provide some technical lemmas and their proofs.
\subsection{Proofs of Lemma~\ref{lemma1} and~\ref{eq:feasible}} 
\label{sec:lemma12}
\begin{proof}[of Lemma~\ref{lemma1}]
For simplicity of notation, we directly use $\mathcal{G}$ and $\mathcal{G}'$ to represent the events $\vz\in\mathcal{G}$ and $\vz'\in\mathcal{G}'$, respectively, when no confusion can be caused. Because $\vz=(\vxi,\zeta,\gamma)$ and $\vz'=(\vxi',\zeta,\gamma)$ are i.i.d. data samples, we have  $\E\big[G_1(\vz)+G_2(\vz') |\mathcal{G}, \mathcal{G}'\big]=\E\big[G_1(\vz)|\mathcal{G}\big]+\E\big[G_2(\vz')|\mathcal{G}'\big]$ and $\E\big[G_1(\vz)G_2(\vz') |\mathcal{G}, \mathcal{G}'\big]=\E\big[G_1(\vz) |\mathcal{G}\big]\E\big[G_2(\vz')|\mathcal{G}'\big]$ for any measurable functions $G_1$ and $G_2$. Based on this fact, we have
\small
\begin{eqnarray}
\nonumber
	&&\E\big[(h_{\vw}(\vxi)-h_{\vw}(\vxi')-c_2)^2|\mathcal{G}, \mathcal{G}'\big]\\\nonumber
 &=&c_2^2 -2c_2\E\big[h_{\vw}(\vxi)|\mathcal{G}\big]+2c_2\E\big[h_{\vw}(\vxi')|\mathcal{G}'\big]
 +\E\big[(h_{\vw}(\vxi))^2|\mathcal{G}\big]+\E\big[(h_{\vw}(\vxi'))^2|\mathcal{G}'\big]
 -2\E\big[h_{\vw}(\vxi)|\mathcal{G}\big]\E\big[h_{\vw}(\vxi')|\mathcal{G}'\big]\\\nonumber
&=&c_2^2-2c_2\E\big[h_{\vw}(\vxi)|\mathcal{G}\big]+2c_2\E\big[h_{\vw}(\vxi')|\mathcal{G}'\big]+\E\big[(h_{\vw}(\vxi))^2|\mathcal{G}\big]-(\E\big[h_{\vw}(\vxi)|\mathcal{G}\big])^2+\E\big[(h_{\vw}(\vxi'))^2|\mathcal{G}'\big]
-(\E\big[h_{\vw}(\vxi')|\mathcal{G}'\big])^2\\\nonumber
&&+(\E\big[h_{\vw}(\vxi)|\mathcal{G}\big])^2+(\E\big[h_{\vw}(\vxi')|\mathcal{G}'\big])^2 -2\E\big[h_{\vw}(\vxi)|\mathcal{G}\big]\E\big[h_{\vw}(\vxi')|\mathcal{G}'\big]\\\nonumber
&=&c_2^2-2c_2\E\big[h_{\vw}(\vxi)|\mathcal{G}\big]+2c_2\E\big[h_{\vw}(\vxi')|\mathcal{G}'\big]+\min_{a}\E\big[(h_{\vw}(\vxi)-a)^2|\mathcal{G}\big]+\min_{b}\E\big[(h_{\vw}(\vxi')-b)^2|\mathcal{G}'\big]\\\nonumber
&&+\max_{\alpha}\left\{2\alpha\E\big[h_{\vw}(\vxi)|\mathcal{G}\big]-2\alpha\E\big[h_{\vw}(\vxi')|\mathcal{G}'\big]-\alpha^2\right\}\\\nonumber
&=&c_2^2-\frac{2c_2\E\big[h_{\vw}(\vxi)\mathbb{I}_{\mathcal{G}}(\vz)\big]}{\text{Pr}(\vz\in\mathcal{G})}+\frac{2c_2\E\big[h_{\vw}(\vxi')\mathbb{I}_{\mathcal{G}'}(\vz')\big]}{\text{Pr}(\vz'\in\mathcal{G}')}+\min_{a}\frac{\E\big[(h_{\vw}(\vxi)-a)^2\mathbb{I}_{\mathcal{G}}(\vz)\big]}{\text{Pr}(\vz\in\mathcal{G})}+\min_{b}\frac{\E\big[(h_{\vw}(\vxi')-b)^2\mathbb{I}_{\mathcal{G}'}(\vz')\big]}{\text{Pr}(\vz'\in\mathcal{G}')}\\\label{eq:Yingderive}
&&+\max_{\alpha}\left\{2\alpha\left(\frac{\E\big[h_{\vw}(\vxi)\mathbb{I}_{\mathcal{G}}(\vz)\big]}{\text{Pr}(\vz\in\mathcal{G})}-\frac{\E\big[h_{\vw}(\vxi')\mathbb{I}_{\mathcal{G}'}(\vz')\big]}{\text{Pr}(\vz'\in\mathcal{G}')}\right)-\alpha^2\right\}.
\end{eqnarray}
\normalsize
Additionally, given any $\vw\in\mathcal{W}$, the optimal value of $a$, $b$ and $\alpha$ are $\E\big[h_{\vw}(\vxi)|\mathcal{G}\big]$, $\E\big[h_{\vw}(\vxi')|\mathcal{G}'\big]$ and $\E\big[h_{\vw}(\vxi)|\mathcal{G}\big]-\E\big[h_{\vw}(\vxi')|\mathcal{G}'\big]$, respectively. By the definition of $\mathcal{I}_{\mathcal{G},\mathcal{G}'}$, we can restrict the decision variables $a$, $b$ and $\alpha$ in $\mathcal{I}_{\mathcal{G},\mathcal{G}'}$ without changing the optimal objective values within \eqref{eq:Yingderive}. The proof is thus completed by multiplying both sides of \eqref{eq:Yingderive} by $c_1$ and observing that $\vxi'$ and $\vz'$ in \eqref{eq:Yingderive} can be replaced by  $\vxi$ and $\vz$ because they are i.i.d. random variables. 

\end{proof} 

\begin{proof}[of Lemma~\ref{eq:feasible}]
By the assumptions of this lemma, there exists $\vw^{\dagger} \in \mathcal{W}$ such that $h_{\vw^{\dagger}}(\vxi)=c$ for any $\bxi$. Let $\vx^{\dagger}$ be a solution in $\mathcal{X}$ whose $\vw$-component equals $\vw^{\dagger}$ and its remaining components are $ a_1^{\dagger}=a_2^{\dagger}=b_1^{\dagger}=b_2^{\dagger}=a_3^{\dagger}=a_4^{\dagger}=b_3^{\dagger}=b_4^{\dagger}=c$. 

By the definitions of $F_{\mathcal{G},\mathcal{G}'}(\vw,a,b;\vz)$ and $G_{\mathcal{G},\mathcal{G}'}(\vw;\vz)$ in \eqref{eq:GFDef}, we have
\small
\begin{equation*}
\begin{split}
    \mathbb{E}[F_{\mathcal{G}_1',\mathcal{G}_1}(\vx^{\dagger};\vz)]&=\mathbb{E}\left[c_1c_2^2-\frac{2c_1c_2c\mathbb{I}_{\mathcal{G}_1'}(\vz)}{Pr(\vz\in\mathcal{G}_1')}+\frac{2c_1c_2c\mathbb{I}_{\mathcal{G}_1}(\vz)}{Pr(\vz\in\mathcal{G}_1)}+\frac{c_1(c-a_1^{\dagger})^2\mathbb{I}_{\mathcal{G}_1'}(\vz)}{Pr(\vz\in\mathcal{G}_1')}+\frac{c_1(c-b_1^{\dagger})^2\mathbb{I}_{\mathcal{G}_1}(\vz)}{Pr(\vz\in\mathcal{G}_1)}\right]=c_1c_2^2,\\
    \mathbb{E}[G_{\mathcal{G}_1',\mathcal{G}_1}(\vw^{\dagger};\vz)]& = \mathbb{E} \left[\frac{c_1c\mathbb{I}_{\mathcal{G}_1'}(\vz)}{Pr(\vz\in\mathcal{G}_1')}-\frac{c_1c\mathbb{I}_{\mathcal{G}'_1}(\vz)}{Pr(\vz\in\mathcal{G}_1)}\right]=0,\\
    \mathbb{E}[F_{\mathcal{G}_2,\mathcal{G}_2'}(\vx^{\dagger};\vz)]&=\mathbb{E} \left[c_1c_2^2-\frac{2c_1c_2c\mathbb{I}_{\mathcal{G}_2}(\vz)}{Pr(\vz\in\mathcal{G}_2)}+\frac{2c_1c_2c\mathbb{I}_{\mathcal{G}_2'}(\vz)}{Pr(\vz\in\mathcal{G}_2')}+\frac{c_1(c-a_2^{\dagger})^2\mathbb{I}_{\mathcal{G}_2}(\vz)}{Pr(\vz\in\mathcal{G}_2)}+\frac{c_1(c-b_2^{\dagger})^2\mathbb{I}_{\mathcal{G}_2'}(\vz)}{Pr(\vz\in\mathcal{G}_2')}\right]=c_1c_2^2,\\
    \mathbb{E}[G_{\mathcal{G}_2,\mathcal{G}_2'}(\vw^{\dagger};\vz)]& =\mathbb{E}\left[ \frac{c_1c\mathbb{I}_{\mathcal{G}_2}(\vz)}{Pr(\vz\in\mathcal{G}_2)}-\frac{c_1c\mathbb{I}_{\mathcal{G}_2'}(\vz)}{Pr(\vz\in\mathcal{G}_2')}\right] = 0.\\
\end{split}
\end{equation*}
\normalsize
Since $c_1c_2^2\leq 0.5$, applying the equations above to the definitions of  $f_1(\vx)$ in \eqref{eq:f1} and \eqref{eq:f2} leads to $f_1(\vx^{\dagger})= 2c_1c_2^2 \leq 1 < 1+\kappa$. Similarly, it holds that $f_2(\vx^{\dagger}) < 1+\kappa$. This means $\vx^{\dagger}$ is a strictly feasible solution and Assumption \ref{assum1} holds.
\end{proof}

\subsection{Closed-Form Solutions for \eqref{eq:BregmanProx} and \eqref{eqn:compUBExpression}} 
\label{sec:closedform}
The closed form of $\vx^{(t+1)}$ is obvious so we only show the closed form of $\vy^{(t+1)}$ in \eqref{eq:BregmanProx}. Given any $\tau>0$, $\bv=(v_0,v_1,v_2,v_3,v_4)\in\mathbb{R}^5$, $\bu=(u_0,u_1,u_2)\in\mathbb{R}^3$ and $\by'=(\tilde\by',\tilde\valpha')\in\mathcal{Y}$, we consider the following problem 
\small
    \begin{align}
   \label{eq:BregmanProx_general}
     \vy^{\#}=(\tilde\by^{\#},\tilde\valpha^{\#})=&\argmin_{\by=(\tilde\by,\tilde\valpha)\in\mathcal{Y}}-(\bu)^\top \tilde \by - (\bv)^\top \tilde \balpha+\frac{V_y(\by,\by')}{\tau}+d_y(\by),
    \end{align}
  \normalsize
which becomes \eqref{eq:BregmanProx} after setting $(\bu,\bv)=\vg_\vy^{(t)}$, $\tau=\tau_t$ and $\by'=\by^{(t)}$. The following lemma characterizes the closed form of $\vy^{\#}$.

\begin{lemma}
	\label{lem:closedform}
Let $  \alpha_0':=\frac{\tilde \alpha_0'}{\tilde y_0'}$, $  \alpha_1':=\frac{\tilde \alpha_1'}{\tilde y_1'}$, $  \alpha_2':=\frac{\tilde \alpha_2'}{\tilde y_1'}$, $  \alpha_3':=\frac{\tilde \alpha_3'}{\tilde y_2'}$, $  \alpha_4':=\frac{\tilde \alpha_4'}{\tilde y_2'}$ and let
\begin{eqnarray}
\label{eq:BregmanProxtildei}
\mu_i:=\min\limits_{ \alpha_i\in\mathcal{I}}\left\{
\begin{array}{l}
- \alpha_i v_i +  \alpha_i^2 + \frac{1}{\tau}\left(  \alpha_i-  \alpha_i'\right)^2
\end{array}\right\}
\quad\text{and}\quad \alpha_i^{\#}:=\argmin\limits_{ \alpha_i\in\mathcal{I}}\left\{
\begin{array}{l}
- \alpha_i v_i +  \alpha_i^2 + \frac{1}{\tau}\left(  \alpha_i-  \alpha_i'\right)^2
\end{array}\right\}.
\end{eqnarray}
for $i=0,1,\dots,4$. Let $\pi_0 := (\tilde y_0')\exp\left(-\frac{\mu_0-u_0 }{2(1+\sqrt{2}I)^2(1/\tau)}\right)$, $\pi_1 := (\tilde y_1')\exp\left(-\frac{\mu_1+\mu_2-u_1 }{2(1+\sqrt{2}I)^2(1/\tau)}\right)$ and $\pi_2 := (\tilde y_2')\exp\left(-\frac{\mu_3+\mu_4-u_2 }{2(1+\sqrt{2}I)^2(1/\tau)}\right)$. Then, $\vy^{\#}=(\tilde \by^{\#}, \tilde \balpha^{\#})\in\mathcal{Y}$ defined as follows is an optimal solution to \eqref{eq:BregmanProx_general}: 
\begin{eqnarray*}
	\label{eq:OptY}
    &&\tilde y_i^{\#} := \frac{\pi_i}{\pi_0+\pi_1+\pi_2} \quad \text{for}\quad i = 0, 1, 2.\\
	&&\tilde \alpha_0^{\#}:=\tilde y_0^{\#} \alpha_0^{\#}, \quad \tilde \alpha_1^{\#}:=\tilde y_1^{\#} \alpha_1^{\#},\quad \tilde \alpha_2^{\#}:=\tilde y_1^{\#} \alpha_2^{\#}, \quad\tilde \alpha_3^{\#}:=\tilde y_2^{\#} \alpha_3^{\#},\quad \tilde \alpha_4^{\#}:=\tilde y_2^{\#} \alpha_4^{\#}.
\end{eqnarray*}
\end{lemma}

\begin{proof}
Recall the definitions of $V_y(\by,\by')$ in \eqref{eq:Distancey} and $d_y(\by)$ in \eqref{eq:dy}.
\eqref{eq:BregmanProx_general} can be formulated as
\begin{eqnarray}
\label{eqn:lemma3_1}
   \min\limits_{\by \in \mathcal{Y}}
   \left\{
   \begin{array}{c}
   -(\bu)^\top \tilde \by - (\bv)^\top \tilde \balpha +\frac{2(1+\sqrt{2}I)^2}{\tau}\sum^2_{i=0}\tilde y_i\ln(\frac{\tilde y_i}{\tilde y_i'})\\
   +\frac{\tilde y_0}{\tau}(\frac{\tilde \alpha_0}{\tilde y_0}-\frac{\tilde \alpha_0'}{\tilde y_0'})^2+\frac{\tilde y_1}{\tau}(\frac{\tilde \alpha_1}{\tilde y_1}-\frac{\tilde \alpha_1'}{\tilde y_1'})^2+\frac{\tilde y_1}{\tau}(\frac{\tilde \alpha_2}{\tilde y_1}-\frac{\tilde \alpha_2'}{\tilde y_1'})^2
   +\frac{\tilde y_2}{\tau}(\frac{\tilde \alpha_3}{\tilde y_2}-\frac{\tilde \alpha_3'}{\tilde y_2'})^2+\frac{\tilde y_2}{\tau}(\frac{\tilde \alpha_4}{\tilde y_2}-\frac{\tilde \alpha_4'}{\tilde y_2'})^2 \\
   + \frac{\tilde \alpha_0^2}{\tilde y_0} + \frac{\tilde \alpha_1^2}{\tilde y_1} + \frac{\tilde \alpha_2^2}{\tilde y_1} + \frac{\tilde \alpha_3^2}{\tilde y_2} + \frac{\tilde \alpha_4^2}{\tilde y_2}
   \end{array}
   \right\}.
\end{eqnarray}
We first fix $\tilde \by \in \triangle_3$ and only optimize $\tilde\balpha$  in \eqref{eqn:lemma3_1} subject to constraints $\tilde\alpha_0 \in\tilde y_0\cdot \mathcal{I}$, $\tilde\alpha_1 \in\tilde y_1\cdot \mathcal{I}$, $\tilde\alpha_2 \in\tilde y_1\cdot \mathcal{I}$, $\tilde\alpha_3\in\tilde y_2\cdot \mathcal{I}$ and $\tilde\alpha_4\in\tilde y_2\cdot \mathcal{I}$. By changing variables using $ \alpha_0:=\frac{\tilde \alpha_0}{\tilde y_0}$, $ \alpha_1:=\frac{\tilde \alpha_1}{\tilde y_1}$, $ \alpha_2:=\frac{\tilde \alpha_2}{\tilde y_1}$, $ \alpha_3:=\frac{\tilde \alpha_3}{\tilde y_2}$, $ \alpha_4:=\frac{\tilde \alpha_4}{\tilde y_2}$ and $ \alpha_0':=\frac{\tilde \alpha_0'}{\tilde y_0'}$, $ \alpha_1':=\frac{\tilde \alpha_1'}{\tilde y_1'}$, $ \alpha_2':=\frac{\tilde \alpha_2'}{\tilde y_1'}$, $ \alpha_3':=\frac{\tilde \alpha_3'}{\tilde y_2'}$, $ \alpha_4':=\frac{\tilde \alpha_4'}{\tilde y_2'}$, \eqref{eqn:lemma3_1} becomes
\begin{align}
   & \min_{\tilde\by\in\triangle_3}
  \left\{\begin{array}{c}
   -(\bu)^\top \tilde \by + \frac{2(1+\sqrt{2}I)^2}{\tau}\sum^2_{i=0}\tilde y_i\ln(\frac{\tilde y_i}{\tilde y_i'}) 
   + \tilde y_0 \min\limits_{  \alpha_0\in \mathcal{I}}\left[- \alpha_0v_0 +\frac{1}{\tau}( \alpha_0- \alpha_0')+\alpha_0^2\right]\\
   + \tilde y_1 \min\limits_{  \alpha_1,   \alpha_2\in \mathcal{I}}\left[\sum\limits_{i=1}^2- \alpha_iv_i +\frac{1}{\tau}( \alpha_i- \alpha_i')+ \alpha_i^2\right] + \tilde y_2 \min\limits_{  \alpha_3,   \alpha_4\in \mathcal{I}}\left[\sum\limits_{i=3}^4- \alpha_iv_i +\frac{1}{\tau}( \alpha_i- \alpha_i')+ \alpha_i^2\right]
  \end{array}\right\}\label{eqn:lemma3_2}\\
  = & \min_{\tilde\by\in\triangle_3}
  \left\{\begin{array}{c}
   -(\bu)^\top \tilde \by + \frac{2(1+\sqrt{2}I)^2}{\tau}\sum^2_{i=0}\tilde y_i\ln(\frac{\tilde y_i}{\tilde y_i'}) + \tilde y_0 \mu_0 + \tilde y_1 (\mu_1+\mu_2) + \tilde y_2 (\mu_3+\mu_4)
  \end{array}\right\},\label{eqn:lemma3_3}
\end{align}
according to the definition of $\mu_i$ in \eqref{eq:BregmanProxtildei}. 

Equality \eqref{eqn:lemma3_2} above indicates that the minimization over $\tilde \valpha$ in \eqref{eqn:lemma3_1} for a given $\tilde \by$ is equivalent to the inner minimization over $\valpha$ in \eqref{eqn:lemma3_2}, which is independent of $\tilde\vy$ and can be solved for each $i$ separately. Note that the optimal objective value and the solution of the $i$th inner minimization are $\mu_i$ and $\alpha_i^{\#}$ in \eqref{eq:BregmanProxtildei}, where $\alpha_i^{\#}$ has a closed form. Equality \eqref{eqn:lemma3_3} indicates that, after obtaining the optimal $\alpha_i$, we can solve the optimal $\tilde \by$ by solving the outer minimization problem \eqref{eqn:lemma3_3} whose solution is exactly $\tilde \by^{\#}$ defined in Lemma \ref{lem:closedform} which can be verified from the optimality conditions. According to the relationship between $ \alpha_i$ and $\tilde \alpha_i$, the optimal value of the original variable $\tilde\alpha_i$ is exactly $\tilde \alpha_i^{\#}$ defined in Lemma \ref{lem:closedform}.

\end{proof}

Next, we consider the optimal value $U(r)$ in \eqref{eqn:compUBExpression}. According to the definition of $\Phi$ in \eqref{eq:Phi}, \eqref{eqn:compUBExpression} can be written as 
\small
    \begin{align}
   \label{eqn:compUBExpression_general}
  U(r) = \max_{\vy=(\tilde\vy,\tilde\valpha) \in \mathcal{Y}} (\vu)^\top \tilde \vy + (\vv)^\top \tilde\valpha - d_y(\vy)+\frac{\hat\rho}{2}\frac{\sum_{t=0}^{T-1}\|\vx^{(t)}-\tilde\vx\|_2^2}{\sum_{t=0}^{T-1}\tau_t},
    \end{align}
\normalsize
where 
\small
$$
\vu=\frac{\sum_{t=0}^{T-1}\vF(\bx^{(t)},\vz^{(t)})}{\sum_{t=0}^{T-1}\tau_t}
\quad\text{and}\quad
\vv= \frac{\sum_{t=0}^{T-1}\vG(\bw^{(t)},\vz^{(t)})}{\sum_{t=0}^{T-1}\tau_t}.
$$  
\normalsize
We denote each component of $\vu$ and $\vv$ as $\bv=(v_0,v_1,v_2,v_3,v_4)\in\mathbb{R}^5$ and $\bu=(u_0,u_1,u_2)\in\mathbb{R}^3$. The following lemma provides a closed form to $U(r)$.

\begin{lemma}
	\label{lem:closedform_u}
 Let $U(r)$ defined in \eqref{eqn:compUBExpression}, or equivalently, in \eqref{eqn:compUBExpression_general}. We have
\begin{eqnarray*}
	\label{eq:OptU}
    U(r) := \max\{u_0+\mu_0, u_1+\mu_1+\mu_2, u_2+\mu_3+
    \mu_4\} + \frac{\hat\rho}{2}\frac{\sum_{t=0}^{T-1}\|\vx^{(t)}-\tilde\vx\|_2^2}{\sum_{t=0}^{T-1}\tau_t},
\end{eqnarray*}
where $\mu_i:=\max\limits_{ \alpha_i\in\mathcal{I}}\left\{\alpha_i v_i -  \alpha_i^2\right\}$ for $i=0,1,\dots,4$.
\end{lemma}
\begin{proof}
Recall the definitions of $d_y(\by)$ in \eqref{eq:dy} and $d_x(\bx)=\frac{\hat\rho}{2}\|\vx-\tilde\vx\|_2^2$. \eqref{eqn:compUBExpression_general} can be formulated as
\small
\begin{eqnarray}
\label{eqn:lemma4_1}
   U(r)=\max\limits_{\by \in \mathcal{Y}}
   \left\{
   \begin{array}{c}
   (\bu)^\top \tilde \by + (\bv)^\top \tilde \balpha  - \frac{\tilde \alpha_0^2}{\tilde y_0} - \frac{\tilde \alpha_1^2}{\tilde y_1} - \frac{\tilde \alpha_2^2}{\tilde y_1} - \frac{\tilde \alpha_3^2}{\tilde y_2} - \frac{\tilde \alpha_4^2}{\tilde y_2}
   \end{array}
   \right\}
   +\frac{\hat\rho}{2}\frac{\sum_{t=0}^{T-1}\|\vx^{(t)}-\tilde\vx\|_2^2}{\sum_{t=0}^{T-1}\tau_t}.
\end{eqnarray}
\normalsize
Similar to the proof of Lemma~\ref{lem:closedform}, we first fix $\tilde \by \in \triangle_3$ and only optimize $\tilde\balpha$  in \eqref{eqn:lemma4_1} subject to constraints $\tilde\alpha_0 \in\tilde y_0\cdot \mathcal{I}$, $\tilde\alpha_1 \in\tilde y_1\cdot \mathcal{I}$, $\tilde\alpha_2 \in\tilde y_1\cdot \mathcal{I}$, $\tilde\alpha_3\in\tilde y_2\cdot \mathcal{I}$ and $\tilde\alpha_4\in\tilde y_2\cdot \mathcal{I}$. By changing variables using $ \alpha_0:=\frac{\tilde \alpha_0}{\tilde y_0}$, $ \alpha_1:=\frac{\tilde \alpha_1}{\tilde y_1}$, $ \alpha_2:=\frac{\tilde \alpha_2}{\tilde y_1}$, $ \alpha_3:=\frac{\tilde \alpha_3}{\tilde y_2}$ and $ \alpha_4:=\frac{\tilde \alpha_4}{\tilde y_2}$, \eqref{eqn:lemma4_1} becomes
\small
\begin{align*}
  U(r)= & \max_{\tilde\by\in\triangle_3}
  \left\{
   (\bu)^\top \tilde \by 
   + \tilde y_0 \max\limits_{  \alpha_0\in \mathcal{I}}\left[\alpha_0v_0 -\alpha_0^2\right]
   + \tilde y_1 \max\limits_{  \alpha_1,   \alpha_2\in \mathcal{I}^2}\left[\sum\limits_{i=1}^2\alpha_iv_i - \alpha_i^2\right]
   + \tilde y_2 \min\limits_{  \alpha_3,   \alpha_4\in \mathcal{I}^2}\left[\sum\limits_{i=3}^4\alpha_iv_i - \alpha_i^2\right]
  \right\}\label{eqn:lemma4_2}\\
  &+ \frac{\hat\rho}{2}\frac{\sum_{t=0}^{T-1}\|\vx^{(t)}-\tilde\vx\|_2^2}{\sum_{t=0}^{T-1}\tau_t}\\
  = & \max_{\tilde\by\in\triangle_3}
  \left\{
   (\bu)^\top \tilde \by + \tilde y_0 \mu_0 + \tilde y_1 (\mu_1+\mu_2) + \tilde y_2 (\mu_3+\mu_4) \right\}
   + \frac{\hat\rho}{2}\frac{\sum_{t=0}^{T-1}\|\vx^{(t)}-\tilde\vx\|_2^2}{\sum_{t=0}^{T-1}\tau_t}\\
  =&\max\{u_0+\mu_0, u_1+\mu_1+\mu_2, u_2+\mu_3+
    \mu_4\} + \frac{\hat\rho}{2}\frac{\sum_{t=0}^{T-1}\|\vx^{(t)}-\tilde\vx\|_2^2}{\sum_{t=0}^{T-1}\tau_t},
\end{align*}
\normalsize
where the second equality is because of the definition of $\mu_i$ for $i=0,\dots,4$ and the last equality is because $\tilde\by\in\triangle_3$. 
\end{proof}

\subsection{A Sufficient Condition for Assumption~\ref{assum:wconvex}.2}
\label{sec:assumption42}
In this subsection, we present the following sufficient condition for  Assumption~\ref{assum:wconvex}.2 to hold. 
\begin{lemma}
Assumption~\ref{assum:wconvex}.2 holds if $2c_1c_2^2-1<\kappa$, $\rho<\frac{2(\kappa+1-2c_1c_2^2)}{\max_{\bx,\bx'\in\mathcal{X}}\|\bx-\bx'\|_2^2}$ and 
there exists $\vw\in\mathcal{W}$ such that $h_{\vw}(\cdot)$ is a constant mapping.
\end{lemma}
\begin{proof}
Because $2c_1c_2^2-1<\kappa$ and $\rho<\frac{2(\kappa+1-2c_1c_2^2)}{\max_{\bx,\bx'\in\mathcal{X}}\|\bx-\bx'\|_2^2}$, there exists $\rho_\epsilon$ such that 
\begin{eqnarray}
\label{eq:rhoepsilon}
0<\rho_\epsilon<\frac{2(\kappa+1-2c_1c_2^2)}{\max_{\bx,\bx'\in\mathcal{X}}\|\bx-\bx'\|_2^2}-\rho.
\end{eqnarray}

Let $\bx\in\mathcal{X}$ be any solution that satisfies $\max_{i=1,2}f_i(\bx)-1-\kappa \leq \epsilon^2$. By the assumptions, there exists $\vw^{\dagger} \in \mathcal{W}$ such that $h_{\vw^{\dagger}}(\vxi)$ is a constant over $\bxi$, denoted by $c$. Let $\vx^{\dagger}$ be a solution in $\mathcal{X}$ whose $\vw$-component equals $\vw^{\dagger}$ and its remaining components are $ a_1^{\dagger}=a_2^{\dagger}=b_1^{\dagger}=b_2^{\dagger}=a_3^{\dagger}=a_4^{\dagger}=b_3^{\dagger}=b_4^{\dagger}=c$. 

According to the proof of Lemma~\ref{eq:feasible} in Section~\ref{sec:lemma12}, we have $f_i(\vx^{\dagger})= 2c_1c_2^2$ for $i=1$ and $2$ and, according to the assumption of this lemma, we have $f_i(\vx^{\dagger})<1+\kappa$ for $i=1$ and $2$. This implies 
 \small
 \begin{eqnarray*}
 &&\min_{\bx'\in\mathcal{X}} \left\{\max_{i=1,2}f_i(\bx') -1-\kappa + \frac{\rho+\rho_\epsilon}{2}\|\bx' - \bx \|_2^2\right\}\\
 &\leq&\max_{i=1,2}f_i(\bx^{\dagger}) -1-\kappa + \frac{\rho+\rho_\epsilon}{2}\|\bx^{\dagger} - \bx \|_2^2\\
 &\leq&2c_1c_2^2-1-\kappa +\frac{\rho+\rho_\epsilon}{2}\max_{\bx,\bx'\in\mathcal{X}}\|\bx-\bx'\|_2^2=-\sigma_\epsilon, 
 \end{eqnarray*}
  \normalsize
where 
$\sigma_\epsilon:=\kappa +1-2c_1c_2^2-\frac{\rho+\rho_\epsilon}{2}\max_{\bx,\bx'\in\mathcal{X}}\|\bx-\bx'\|_2^2$ is a positive number because of \eqref{eq:rhoepsilon}. This completes the proof. 
\end{proof}
\begin{remark}
Condition $2c_1c_2^2-1<\kappa$ means the targeted fairness level should not be too small, so there exists a sufficiently feasible solution (see  Lemma~\ref{eq:feasible}). Condition  $\rho<\frac{2(\kappa+1-2c_1c_2^2)}{\max_{\bx,\bx'\in\mathcal{X}}\|\bx-\bx'\|_2^2}$ means the original non-convex problem should not have a high level of non-convexity.
\end{remark}


\section{DISCUSSION ON THEOREM~\ref{thm:convexcase}}
\label{sec:mainproof}

In this section, we briefly discuss how to directly apply the results from~\cite{nemirovski2009robust,lin2020data} to obtain Theorem~\ref{thm:convexcase}. First, we match our notation to those used in \cite{lin2020data} and instantize the convergence results in \cite{lin2020data} on \eqref{eq:minmax_new}. Recall that $\|\vx\|_x=\|\vx\|_2$ and $\|\by\|_{y}=\sqrt{\|\tilde\by\|_1^2+\|\tilde\valpha\|_2^2}$. Their dual norms are $\|\vx\|_{*,x}=\|\vx\|_2$ and $\|\by\|_{*,y}=\sqrt{\|\tilde\by\|_\infty^2+\|\tilde\valpha\|_2^2}$, respectively. 
The complexity of SMD is known to depend on the diameters of $\mathcal{X}$ and $\mathcal{Y}$ measured by the corresponding distance generating functions, namely, 
$$
D_x:=\sqrt{\max_{\bx\in\mathcal{X}}\omega_x(\bx)-\min_{\bx\in\mathcal{X}}\omega_x(\bx)}\text{   and   }D_y  := \sqrt{\max_{\by\in\mathcal{Y}}\omega_y(\by)-\min_{\by\in\mathcal{Y}}\omega_y(\by)}
$$
defined in Theorem~\ref{thm:convexcase}. Moreover, thanks to Assumption~\ref{assum:deviation}, it is not hard to show that there exist constants $M_x$, $M_y$ and $Q$, which only depend on $\sigma$, $I$, $\rho$ and $D_x$, such that 
	\begin{align}
	\E\left[\exp(\|\nabla_x\Phi(\bx,\by,\vz)\|_{*,x}^2/M_x^2)\right]&\leq\exp(1),\\
	\E\left[\exp(\|\nabla_y\Phi(\bx,\by,\vz)\|_{*,y}^2/M_y^2)\right]&\leq\exp(1),\\
	\E\left[\exp(\left|\Phi(\bx,\by,\vz)-\phi(\bx,\by)\right|^2/Q^2)\right]&\leq\exp(1),
	\end{align}
Additionally, given $\delta\in(0,1)$, we define 
\begin{align}
\label{eq:defM}
M &:=\sqrt{2D_x^2M_x^2+2D_y^2M_y^2};\\
\Omega(\delta) &:= \max\left\{\sqrt{12\ln\left(\dfrac{24}{\delta}\right)},\dfrac{4}{3}\ln\left(\dfrac{24}{\delta}\right)\right\}.\label{eq:defOmega}
\end{align}
With those notations, a brief proof of Theorem~\ref{thm:convexcase} is given below.

Compared with problem (5) in  \cite{lin2020data}, our problem~\eqref{eq:minmax_new} has the additional terms $d_x(\bx)$ and $d_y(\by)$. However, since we choose the initial solution as $\bx^{(0)}=\mathbf{0}$ and $\by^{(0)}=(1/3,1/3,1/3,\mathbf{0})$, these additional terms can be eliminated from the proof of any theorems and propositions in \cite{lin2020data}, so the convergence results in \cite{lin2020data} also hold for problem~\eqref{eq:minmax_new} and Algorithm~\ref{alg:smd}. Moreover, the algorithm in \cite{lin2020data} is presented using a unified updating scheme for $\bx$ and $\by$ with only one step size $\gamma_t$ while our  Algorithm~\ref{alg:smd} is presented with $\bx$ and $\by$ updated separately. However, it is easy to verify that, by choosing $\eta_t=2D_x^2\gamma_t$ and $\tau_t=2D_x^2\gamma_t$ with  $\gamma_t = 1/(M\sqrt{t+1})$ where $M$ is defined in \eqref{eq:defM}, Algorithm~\ref{alg:smd} is equivalent to the algorithm in \cite{lin2020data}. Hence, according to Theorem 8 in \cite{lin2020data}, if
	\small
	\begin{equation}
	\label{eq:TvalueH}
T\geq 	T(\delta,\epsoracle) :=\max\left\{6,\left(\frac{16\left( Q\Omega(\delta)+10 M \Omega(\delta)+4.5M\right)}{\epsoracle}\ln\left(\frac{8\left( Q \Omega(\delta)+10 M \Omega(\delta)+4.5M\right)}{\epsoracle}\right)\right)^2-2\right\},
	\end{equation}
	\normalsize
the outputs $U(r)$ and $\bar \bx$ by Algorithm~\ref{alg:smd} satisfy the inequalities $\Pc(r,\bar \bx)- H(r)\leq\epsilon$ and $|U(r)-H(r)| \leq \epsilon$ with a probability of at least $1-\delta$ for any $r>f^*$. Hence, 
SMD with $T\geq T(\delta,\epsoracle)$ is a valid stochastic oracle  defined in Definition~\ref{def:feasOracle}. Hence, according to Corollary 9 in \cite{lin2020data}, SFLS returns a relative $\epsilon$-optimal and feasible solution with probability of at least $1-\delta$ using at most $\tilde O\left(\frac{1}{\epsilon^2}\ln(\frac{1}{\delta})\right)$ stochastic mirror descent steps across all calls of SMD.  Theorem~\ref{thm:convexcase} is thus proved.

\section{DEFINITION of $\tilde{\phi}$ IN \eqref{eq:oldphi} AND TABLE OF NOTATIONS}
\label{def_formula}
In Section~\ref{sec:convexcase}, we can write \eqref{eq:gcols} as 
$$
H(r):=\min_{\bx\in\mathcal{X}}\Pc(r,\bx)=\min_{\bx\in\mathcal{X}}\max_{\tilde\by\in\Delta_3}\left\{
\tilde y_0(f_0(\bx)-r)+\tilde y_1(f_1(\bx)-1-\kappa)+\tilde y_2(f_2(\bx)-1-\kappa)\right\},
$$
where $\Delta_3:=\{\tilde\by\in\mathbb{R}^3_+|\sum_{i=0}^2\tilde y_i=1\}$. With \eqref{eq:f0}, \eqref{eq:f1} and \eqref{eq:f2}, we can reformulate the problem above into \eqref{eq:oldphi}, i.e., 
\small
\begin{equation*}
  H(r):=  \min_{\bx\in\mathcal{X}}
\max_{\tilde\by\in\Delta_3, \valpha\in \mathcal{I}^5}\tilde{\phi}(\vx,\tilde{\vy},\valpha),
\end{equation*}
\normalsize
where
\small
\begin{eqnarray*}
\label{eq:gcolsminmax}
\tilde{\phi}(\vx,\tilde{\vy},\valpha):=\E\left\{
\begin{array}{c}
-r\tilde y_0-(1+\kappa)\tilde y_1-(1+\kappa)\tilde y_2\\
+\tilde y_0F_{\mathcal{D}_+,\mathcal{D}_-}(\vx;\vz)
+\tilde y_1F_{\mathcal{G}_1',\mathcal{G}_1}(\vx;\vz)+\tilde y_1F_{\mathcal{G}_2,\mathcal{G}_2'}(\vx;\vz)
+\tilde y_2F_{\mathcal{G}_2',\mathcal{G}_2}(\vx;\vz)+\tilde y_2F_{\mathcal{G}_1,\mathcal{G}_1'}(\vx;\vz)\\
+\tilde y_0\alpha_0 G_{\mathcal{D}_+,\mathcal{D}_-}(\vw;\vz)
+\tilde y_1\alpha_1 G_{\mathcal{G}_1',\mathcal{G}_1}(\vw;\vz)+\tilde y_1\alpha_2 G_{\mathcal{G}_2,\mathcal{G}_2'}(\vw;\vz)
+\tilde y_2\alpha_3 G_{\mathcal{G}_2',\mathcal{G}_2}(\vw;\vz)+\tilde y_2\alpha_4 G_{\mathcal{G}_1,\mathcal{G}_1'}(\vw;\vz)\\
-\tilde y_0\alpha_0^2-\tilde y_1\alpha_1^2-\tilde y_1\alpha_2^2-\tilde y_2\alpha_3^2-\tilde y_2\alpha_4^2
\end{array}
\right\}.
\end{eqnarray*}
\normalsize

Since many notations are introduced this paper, we summarize them in Table~\ref{tbl:notation} so readers can find their meanings more easily.  
\begin{table}[ht]
\caption{Notation used throughout the paper.}
\label{tbl:notation}
\vskip 0.1in
\centering
\begin{tabular}{c|c}
\toprule
 Symbol & Definition \\
\midrule
$\vxi$  & Feature vector of a data point. \\
$\zeta$ & Binary label of a  data point.  \\
$\gamma$ & Binary sensitive feature of a  data point. \\
$\vz=(\vxi,\zeta,\gamma)$ & A data point.\\
$\vw$ and $\mathcal{W}$ & Parameters of a classification model. It belongs to a convex compact set $\in\mathcal{W}$.\\
$h_{\vw}(\vxi)$ & Predicted score for a data point based its feature $\vxi$. \\
$\mathcal{G}$, $\mathcal{G}_1$, $\mathcal{G}_1'$, $\mathcal{G}_2$, $\mathcal{G}_2'$   & Set in $\mathbb{R}^{p+2}$ with positive measures w.r.t. $\vz$.\\
$\mathcal{D}_+$ & Positive dataset.\\
$\mathcal{D}_-$ & Negative dataset.\\
$\ell(\cdot)$ & Surrogate loss function that approximates $\mathbb{I}_{(\cdot\leq 0)}$ and $\mathbb{I}_{(\cdot< 0)}$.\\
$c_1(\cdot-c_2)^2$ & Quadratic loss function that approximates $\mathbb{I}_{(\cdot\leq 0)}$ and $\mathbb{I}_{(\cdot< 0)}$.\\
$a,b,\alpha$ & Auxiliary variables introduced to formulate the quadratic loss into a min-max problem~\eqref{eq:minmax}.\\
$\mathcal{I}_{\mathcal{G},\mathcal{G}'}$& The smallest interval that contains
$\left\{\begin{array}{c}
0, \pm\E\big[h_{\vw}(\vxi)|\vz\in\mathcal{G}\big],\pm\E\big[h_{\vw}(\vxi')|\vz'\in\mathcal{G}'\big],\\
\pm\left(\E\big[h_{\vw}(\vxi)|\vz\in\mathcal{G}\big]-\E\big[h_{\vw}(\vxi')|\vz'\in\mathcal{G}'\big]\right)
\end{array}\right\}$.\\
$\mathcal{I}$ and $I$&A bounded interval containing $\mathcal{I}_{\mathcal{D}_+, \mathcal{D}_-},\mathcal{I}_{\mathcal{G}_1, \mathcal{G}_1'},\mathcal{I}_{\mathcal{G}_2,\mathcal{G}_2'}$ and $I:=\max_{\alpha\in\mathcal{I}}|\alpha|$.\\
$\mathcal{X}$&The domain of primal variables.\\ 
$\mathcal{Y}$&The domain of dual variables.\\
$\Delta_3$& The simplex in $\mathbb{R}^3$.\\
$\omega_x(\vx)$ and $\omega_y(\vx)$ & Distance generating functions on $\mathcal{X}$ and $\mathcal{Y}$, respectively.\\
$V_x(\bx,\bx')$ and $V_y(\by,\by')$ & Bregman divergences on $\mathcal{X}$ and $\mathcal{Y}$, respectively. \\
$H(r)$ and $\widehat H(r)$&Level-set functions of \eqref{eq:minmaxproblem} and \eqref{eq:phix}, respectively. \\
$r$ and $r^{(k)}$& Level parameters in the stochastic level-set method.\\
$\rho$ and $\hat\rho$& Weak convexity parameter of  \eqref{eq:minmaxproblem} and  $\hat\rho>\rho$.\\
\bottomrule
\end{tabular}
\end{table}

\section{ADDITIONAL MATERIALS FOR NUMERICAL EXPERIMENTS}
In this section, we present some additional details of our numerical experiments in Section~\ref{sec:exp}. 

\subsection{Details of Datasets}
\label{sec:data}
We provide below some details about the three datasets we used in our numerical experiments.
\begin{itemize}
    \item The \textit{a9a} dataset is used to predict if the annual income of an individual exceeds \$50K. Gender is  the sensitive attribute, i.e., female ($\gamma=1$) or male ($\gamma=-1$).
    \vspace{-0.05in}
    \item The \textit{bank} dataset is used to predict if a client will subscribe a term deposit. Age is the sensitive attribute, i.e., age between 25 and 60 ($\gamma=1$) or otherwise ($\gamma=-1$).
    \vspace{-0.05in}
    \item The \textit{COMPAS} dataset is used to predict if a criminal defendant will reoffend. Race is the sensitive attribute, i.e., caucasian ($\gamma=1$) or non-caucasian ($\gamma=-1$).
\end{itemize}
Some statistics of these datasets are given in Table \ref{tbl:data}. Data \textit{a9a} originally has a training set and a testing set, and we further split the training data into a training set (\%90) and a validation set (\%90). For \textit{bank} and \textit{COMPAS} datasets, we split them into training (\%60), validation (\%20) and testing (\%20) sets. The validation sets are used for tuning hyper-parameters while the testing sets are for performance evaluation.

\begin{table}[ht]
\caption{Statistics of the datasets.}
\label{tbl:data}
\vskip 0.1in
\centering
\begin{tabular}{c|ccccc}
\toprule
 Datasets & \#Instances & \#Attributes & Class Label & Sensitive Attribute \\
\midrule
a9a & 48,842 & 123 & Income & Gender\\
bank & 41,188 & 54 & Subscription & Age\\
COMPAS & 11,757 & 14 & Recidivism & Race\\
\bottomrule
\end{tabular}
\end{table}

\subsection{Details of Baselines}
\label{sec:baseline}
In this section, we provide the details of three baselines used in our experiments.

\begin{itemize}
    \item Proxy-Lagrangian is a Lagrangian method for solving \eqref{eqn:problem}, where only the indicator function $\mathbb{I}_{(h_{\vw}(\vxi)-h_{\vw}(\vxi')\leq 0)}$ in the objective function is approximated by a surrogate loss while the indicator functions in the constraints are unchanged. 
    \item Correlation-penalty is a method that adds the absolute value of the correlation between $h_{\vw}(\vxi)$ and $\gamma$ in the objective function as a penalty term while optimizing the AUC of $h_{\vw}$ for predicting $\zeta$. We are only able to apply this method when the fairness constraints are based on Example \ref{ex:GroupAUC} because the constraints based on  Examples \ref{ex:xAUC} and \ref{ex:intraAUC} cannot be equivalently represented as penalty terms of statistical correlations.
    \item In the post-processing method, we first train a model by optimizing the AUC of $h_{\vw}$ for predicting $\zeta$ without any constraints. Then we modify the predicted scores on data with $\gamma=1$ to $\omega_1 h_{\vw}(\vxi)+\omega_2$ but leave the scores on data with $\gamma=-1$ unchanged. We then tune $\omega_1$ and $\omega_2$ to satisfy the constraints in \eqref{eqn:problem}. We are unable to apply post-processing to Example \ref{ex:intraAUC} since tuning $\omega_1$ and $\omega_2$ requires knowing the true labels ($\zeta$) of the data, which is impractical.
\end{itemize}

\subsection{Process of Tuning Hyperparameters}
\label{sec:tune}
In this section, we explain the process to tune the hyper-parameters.

\textbf{Convex case.} For the level-set method and the proxy-Lagrangian method, we solve their constrained optimization problems with different values of $\kappa$. For each value of $\kappa$, we track the models from all iterations and return the one that is feasible to \eqref{eqn:problem} and reaches the best AUC on the validation set. In the correlation-penalty method, we select $\lambda$ from a set of candidates, solve the penalized optimization problem by the stochastic gradient descent method, and select the model to return in the same way as the previous two methods. We set $c_2 = 1$ and choose $c_1$ from 0.5 and 1 for all methods. For the level-set method, we set $\theta=1$ in Algorithm \ref{SFLS}  and  $\eta_t=\frac{c}{\sqrt{t+1}}$ in Algorithm \ref{alg:smd} with $c$ tuned from $\{10^{-2},10^{-1},1\}$ based on the AUC of the returned model on the validation set. The learning rates of proxy-Lagrangian and correlation-penalty are tuned in the same way. For post-processing, $\omega_1$ is tuned from a grid in $[0, 5]$ with a gap of 0.05 and $\omega_2$ is tuned from a grid in  $[-3, 3]$ with step size 0.1. We use a mini-batch of size 100 in each method when computing stochastic gradients.

\textbf{Weakly-convex case.} The implementation of each method and the process of tuning hyperparameters is the same as the convex case except that we choose $\hat{\rho}=10^{-5}$ in Algorithm~\ref{alg:iqrc}.

\subsection{Plots of COMPAS Dataset}
\label{sec:compas}
\begin{figure}
    \centering
    \includegraphics[width=0.33\textwidth]{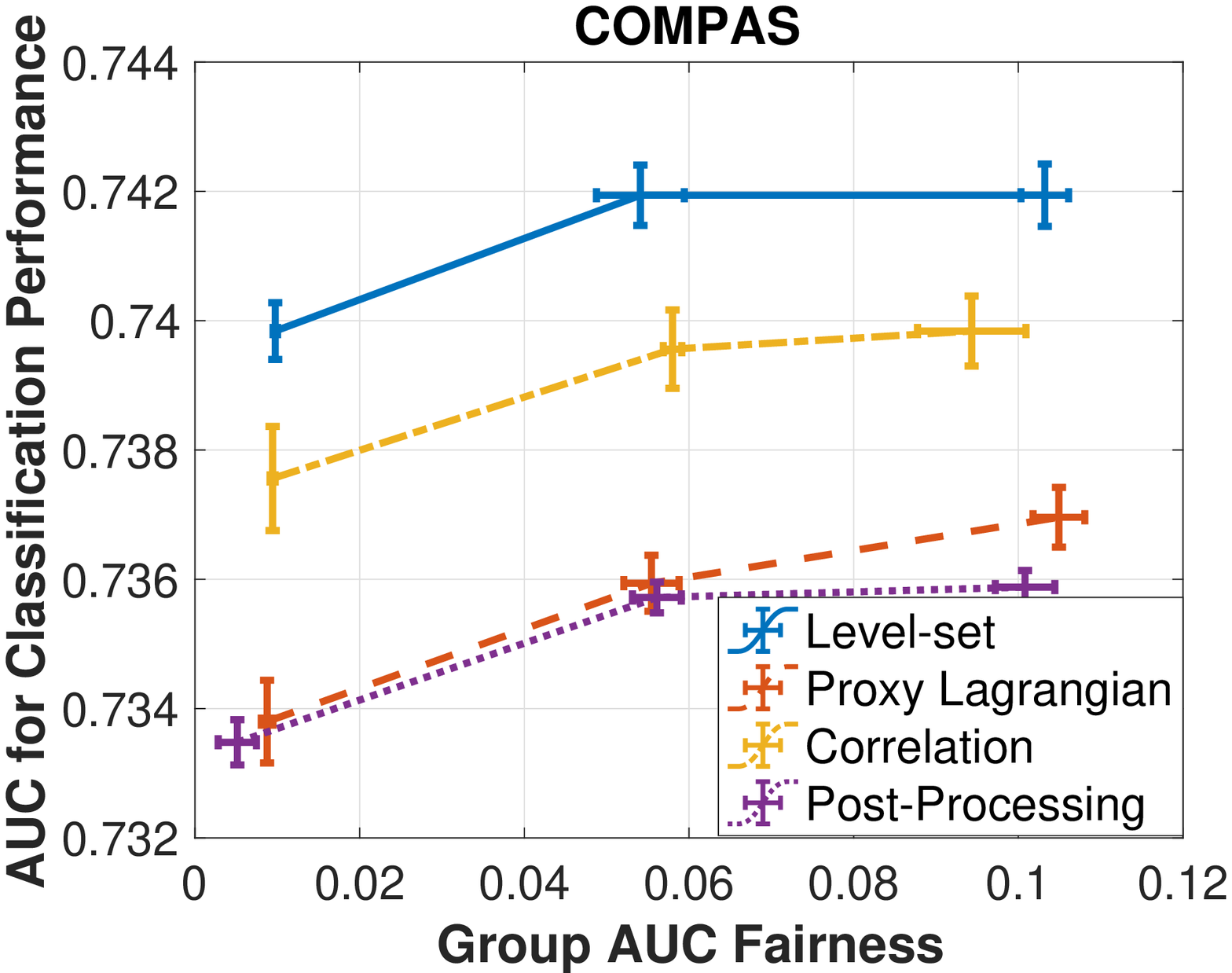}\vspace*{-0.01in}
    \includegraphics[width=0.33\textwidth]{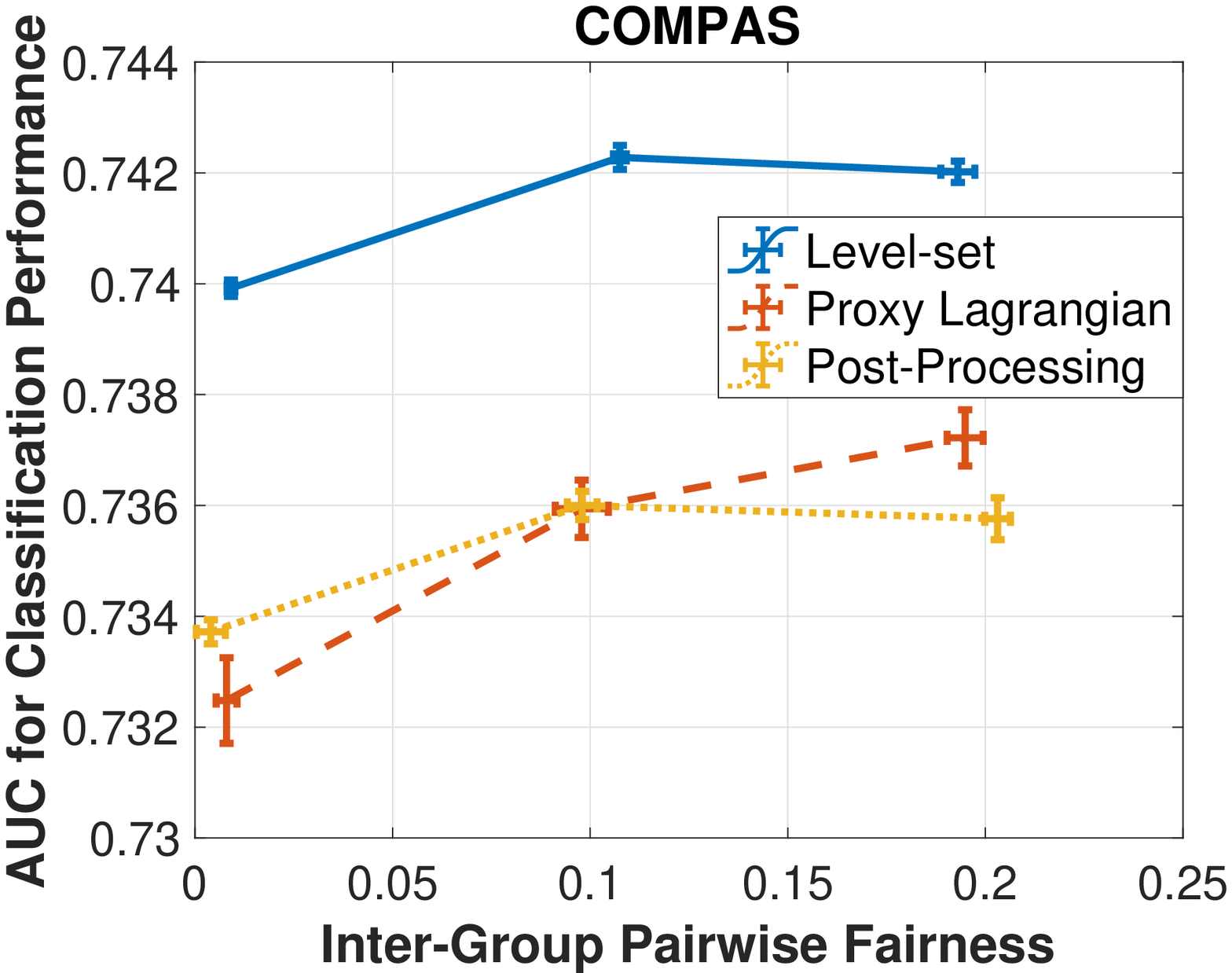}\vspace*{-0.01in}
    \includegraphics[width=0.33\textwidth]{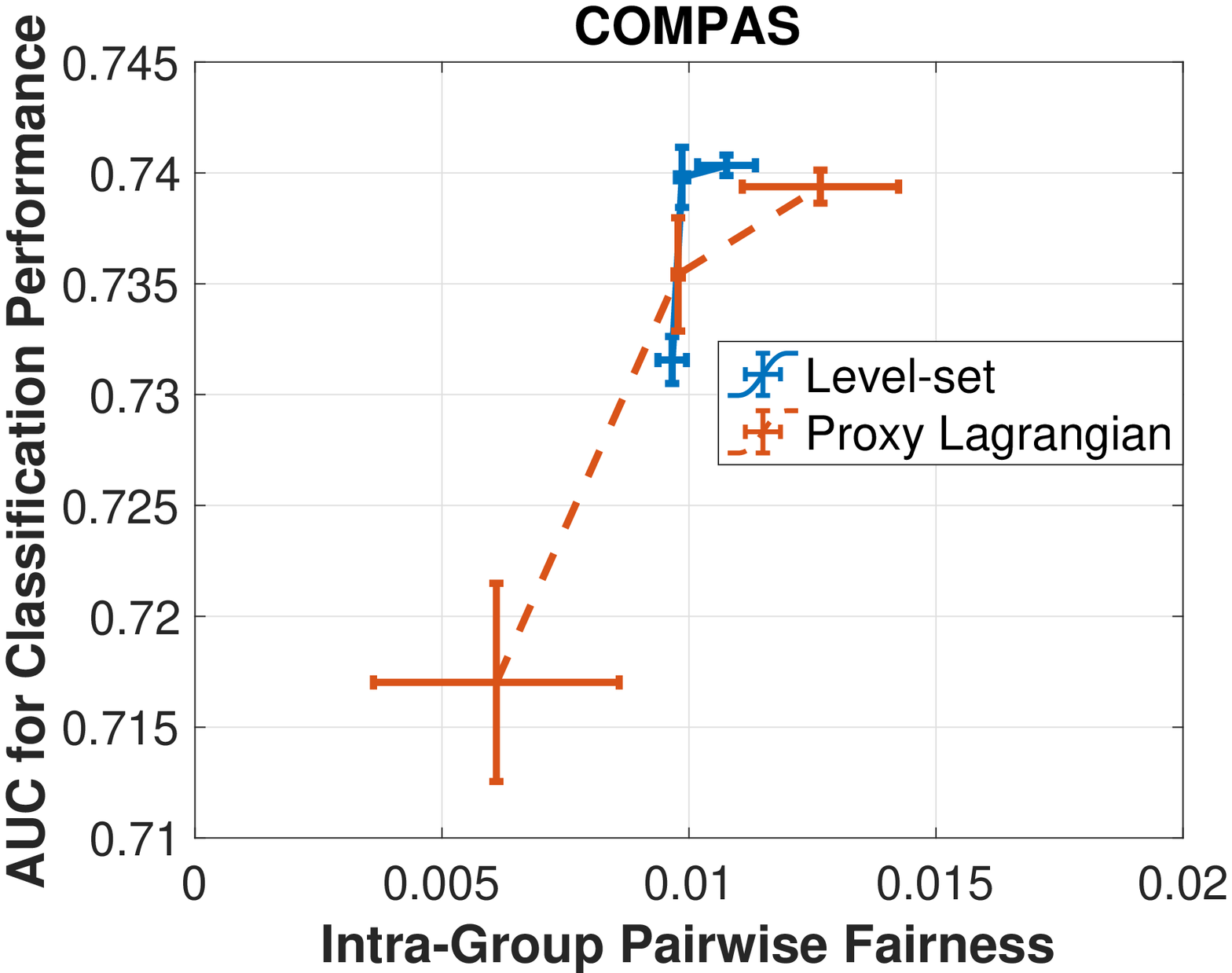}\vspace*{-0.01in}
    \caption{Pareto frontiers by each method for COMPAS dataset in convex case.}
    \label{fig:COMPAS}
\end{figure}

\begin{figure}
    \centering
    \includegraphics[width=0.33\textwidth]{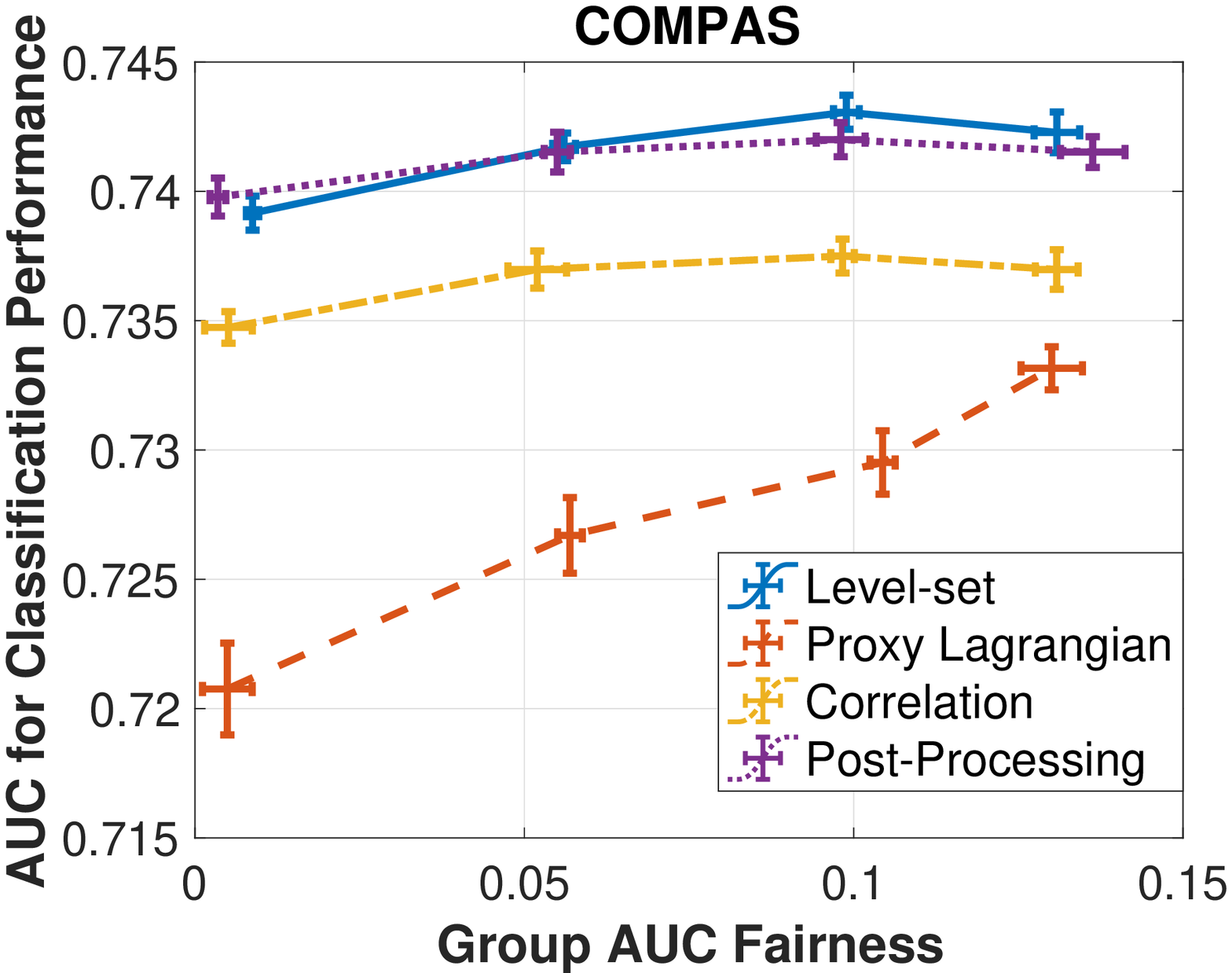}\vspace*{-0.01in}
    \includegraphics[width=0.33\textwidth]{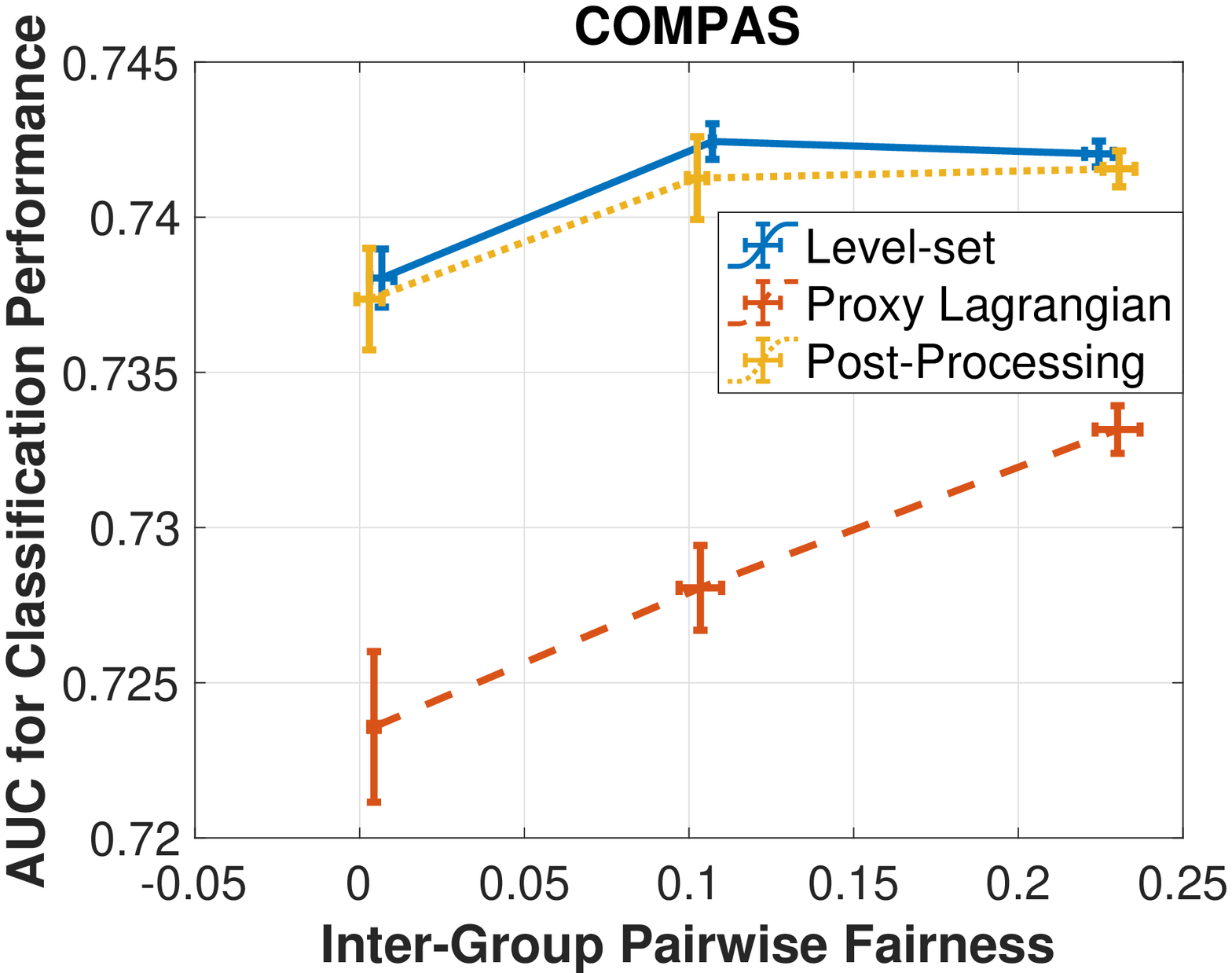}\vspace*{-0.01in}
    \includegraphics[width=0.33\textwidth]{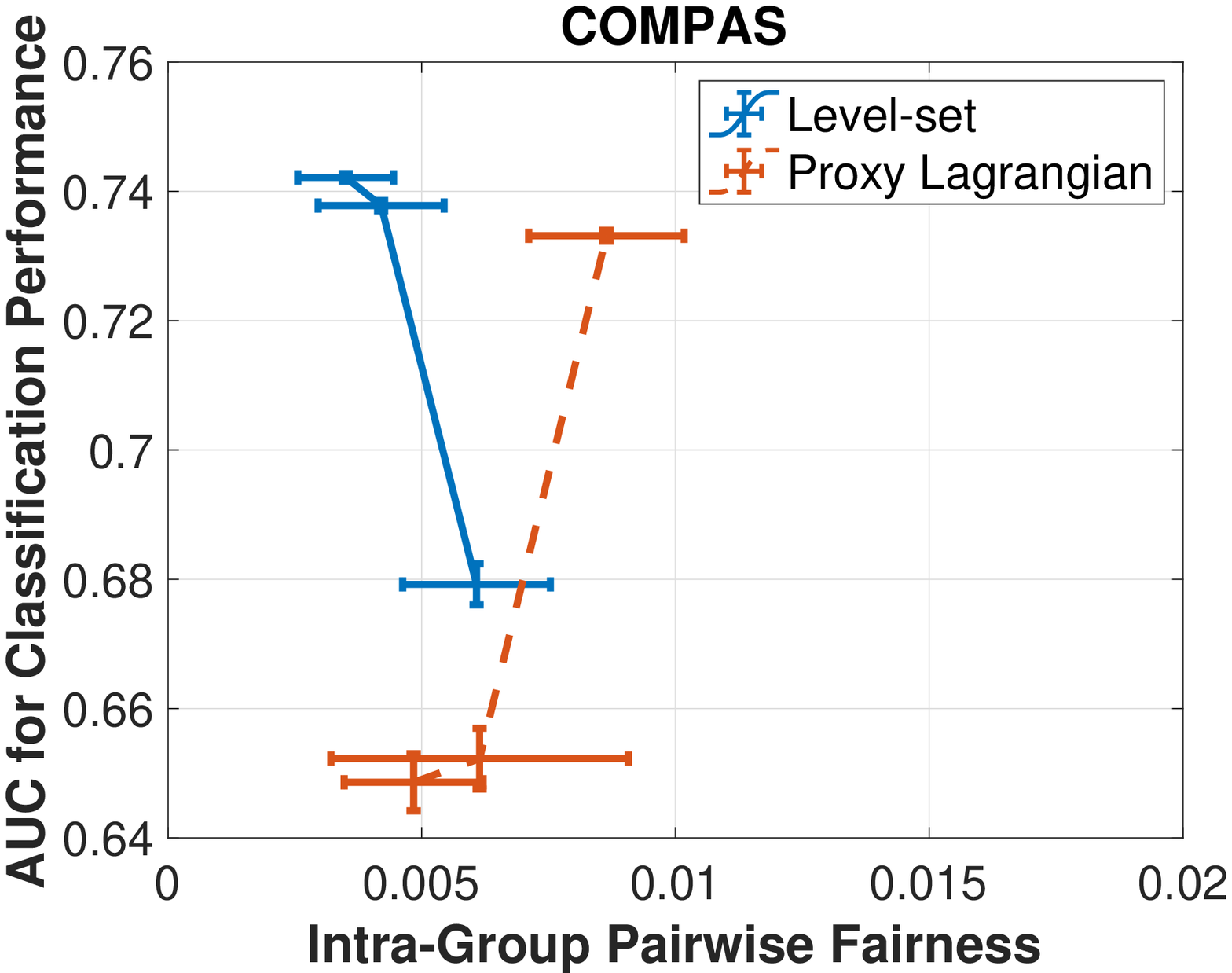}\vspace*{-0.01in}
    \caption{Pareto frontiers by each method for COMPAS dataset in weakly-convex case.}
    \label{fig:COMPAS_nonconvex}
\end{figure}

In this section, we present the Pareto frontier obtained by each method on the COMPAS dataset in Figures~\ref{fig:COMPAS} and \ref{fig:COMPAS_nonconvex} for the convex case and the weakly-convex case, respectively. 

\end{document}